\pdfoutput=1

\documentclass[11pt]{article}

\usepackage[final]{acl}

\usepackage{times}
\usepackage{latexsym}

\usepackage[T1]{fontenc}

\usepackage[utf8]{inputenc}

\usepackage{microtype}

\usepackage{inconsolata}

\usepackage{graphicx}
\usepackage{multirow}
\usepackage{acl}
\usepackage{xcolor}
\usepackage{colortbl}
\usepackage{hyperref}
\usepackage{url}
\usepackage{graphicx} 
\usepackage{longtable}
\usepackage{diagbox}
\usepackage{array}
\usepackage{tabularx} 
\usepackage{booktabs}
\usepackage{amsmath} 
\usepackage{amssymb}
\usepackage{algorithm,algorithmic}
\usepackage{helvet} 
\usepackage{subcaption} 
\usepackage{caption}
\usepackage[hypcap=true]{caption}
\title{Large Language Model Evaluation via Matrix Nuclear-Norm}

\author{
   Yahan Li\textsuperscript{\rm 1}, Tingyu Xia\textsuperscript{\rm 1}, Yi Chang\textsuperscript{\rm 1,\rm 2,\rm 3}, Yuan Wu\textsuperscript{\rm 1,\rm 2}\thanks{Corresponding author} \\
   \textsuperscript{\rm 1}School of Artificial Intelligence, Jilin University\\
   \textsuperscript{\rm 2}Key Laboratory of Symbolic Computation and Knowledge Engineering, Jilin University\\
   \textsuperscript{\rm 3}International Center of Future Science, Jilin University\\
   \texttt{yahan23@mails.jlu.edu.cn, xiaty21@mails.jlu.edu.cn, yichang@jlu.edu.cn} \\
   \texttt{yuanwu@jlu.edu.cn} \\
 }

\begin{document}

\maketitle

\begin{abstract}

As large language models (LLMs) continue to evolve, efficient evaluation metrics are vital for assessing their ability to compress information and reduce redundancy. While traditional metrics like Matrix Entropy offer valuable insights, they are computationally intensive for large-scale models due to their \( O(n^3) \) time complexity with Singular Value Decomposition (SVD). To mitigate this issue, we introduce the Matrix Nuclear-Norm, which not only serves as a metric to quantify the data compression proficiency of LLM but also provides a convex approximation of matrix rank to capture both predictive discriminability and diversity. By employing the \( L_{1,2}\text{-norm} \) to further approximate the nuclear norm, we can effectively assess the model's information compression capabilities. This approach reduces the time complexity to \( O(n^2) \) and eliminates the need for SVD computation. Consequently, the Matrix Nuclear-Norm achieves speeds 8 to 24 times faster than Matrix Entropy for the Cerebras-GPT model as sizes increase from 111M to 6.7B. This performance gap becomes more pronounced with larger models, as validated in tests with other models like Pythia. Additionally, evaluations on benchmarks and model responses confirm that our proposed Matrix Nuclear-Norm is a reliable, scalable, and efficient tool for assessing LLMs' performance, striking a balance between accuracy and computational efficiency. 
The code is available at \url{https://github.com/MLGroupJLU/MatrixNuclearNorm}.

\end{abstract}

\section{Introduction}
Large language models (LLMs), such as Gemini~\citep{team2023gemini}, Deepseek~\citep{guo2025deepseek}, and GPT-4~\citep{achiam2023gpt}, have shown exceptional performance in numerous natural language processing (NLP) tasks~\citep{zhao2023survey}. These models are transforming the way we approach NLP tasks, providing unprecedented capabilities and solutions to complex problems. They are revolutionizing NLP~\citep{saul2005advances, liu2023llmrec, sawada2023arb} and positively impacting computer vision \citep{lian2023llm, wang2024visionllm} and graph neural networks \citep{zhang2024benchmarking, chen2024exploring}, achieving top results on leaderboards. Despite these advancements, evaluating a model's ability to compress information remains a critical research challenge \citep{deletang2023language}. This challenge is essential for improving the overall efficiency of these models.

Compression involves efficiently extracting essential information from large datasets while removing redundant data, highlighting a model’s ability to understand the data's underlying structure~\citep{wei2024large}. LLMs are expected to perform this compression during training~\citep{zhao2023survey}. Initially, after random initialization, the data representations are chaotic, but as training progresses, they become organized, allowing the model to filter out unnecessary information. Thus, assessing an LLM’s compression capacity is vital for understanding its learning efficiency and representational power, which are crucial for practical applications and real-world deployment.

Current compression metrics like \citet{wei2024large}'s Matrix Entropy analyze output representations but face scalability limits due to $O(n^3)$ SVD complexity \citep{kung1983state,zhang2015singular}. 
To address this, we propose a novel metric called Matrix Nuclear-Norm. This metric measures predictive discriminability and output diversity, serving as an upper bound for the Frobenius norm and providing a convex approximation of the matrix rank. We enhance the Matrix Nuclear-Norm by using the \(L_{1,2}\text{-norm}\) to approximate the nuclear norm, improving stability across multiple classes. This approach efficiently assesses a model's compression capabilities and redundancy elimination, streamlining evaluation. The Matrix Nuclear-Norm has a computational complexity of \(O(n^2)\), a significant improvement over Matrix Entropy’s \(O(n^3)\). This optimization achieves $>8\times$ acceleration in evaluation speed for large models while preserving reliability.

To validate the Matrix Nuclear-Norm, we conducted preliminary experiments on two language models of different sizes. Results showed a consistent decrease in Matrix Nuclear-Norm values as model size increased, indicating enhanced compression capabilities. We also performed inference experiments on benchmark datasets, AlpacaEval \citep{dubois2024length} and Chatbot Arena \citep{chiang2403chatbot}, covering diverse language generation tasks. These benchmarks provide a comprehensive assessment of model inference performance. Our findings confirm that the Matrix Nuclear-Norm accurately measures model compression capabilities and ranks models based on performance, demonstrating its reliability and efficiency. Our empirical investigations yield the following insights:

\begin{itemize}
    \setlength{\labelsep}{0.4em} 
    \setlength{\leftmargin}{10em} 
    \setlength{\itemsep}{1pt} 
    \setlength{\parskip}{1pt} 
    \item \textbf{Proposal of the Matrix Nuclear-Norm}: We introduce a method leveraging the nuclear norm, reducing computational complexity from \(O(n^3)\) to \(O(n^2)\). This reduction minimizes SVD dependence, making Matrix Nuclear-Norm a more efficient alternative to Matrix Entropy.
    \item \textbf{Extensive Experimental Validation}: We validated the Matrix Nuclear-Norm on language models of various sizes. Results show this metric accurately assesses model compression capabilities, with values decreasing as model size increases, reflecting its robust evaluation capability.
    \item \textbf{Benchmark Testing and Ranking}: We conducted inference tests on benchmark datasets, AlpacaEval and Chatbot Arena, evaluating inference performance across different model sizes and ranking them based on the Matrix Nuclear-Norm. Results demonstrate this metric efficiently and accurately evaluates medium and small-scale models, highlighting its broad application potential in model performance assessment.
\end{itemize}

\section{Related Work}

\textbf{LLM Evaluation and Scaling Laws.}  
Evaluating large language models (LLMs) is a multifaceted challenge, as it requires capturing both task-specific performance and internal representational efficiency. Scaling laws have become a foundational framework for studying how LLM performance evolves with model size and data volume \citep{kaplan2020scaling, ruan2024observational}. These studies demonstrate that model performance on tasks like language modeling and fine-tuning often follows predictable power-law relationships with respect to model parameters and dataset size, emphasizing the importance of scaling for achieving state-of-the-art results.However, scaling laws typically focus on external metrics such as cross-entropy loss, offering limited insight into how LLMs manage internal knowledge representation. For instance, the ability of LLMs to compress knowledge, eliminate redundancy, and retain structured information remains poorly understood with traditional methods. Addressing these gaps requires structural metrics that go beyond task outcomes to directly evaluate the internal embeddings and activation patterns of LLMs.

\textbf{LLM Evaluation Metrics.}  
Traditional evaluation metrics such as perplexity, BLEU \citep{papineni2002bleu}, and ROUGE \citep{lin2004rouge} primarily measure task-specific outcomes, assessing how well model outputs align with ground truth data. While these metrics are effective for evaluating surface-level outputs, they do not capture the underlying mechanisms of LLMs, such as the diversity or compression of embeddings. Similarly, accuracy and F1 score \citep{2007The} focus on classification performance, making them less applicable to the generative tasks typical of LLMs.To bridge this gap, structural metrics such as Matrix Entropy have been introduced. Matrix Entropy \citep{wei2024large} employs information theory to assess the entropy of covariance matrices derived from LLM embeddings. This metric evaluates how effectively a model removes redundancy and encodes structured information, offering a measure of its compression capabilities. For instance, Matrix Entropy can reveal differences in embedding distributions across models of varying sizes, reflecting their capacity to extract meaningful patterns from large datasets. However, its reliance on Singular Value Decomposition (SVD) results in a computational complexity of $O(n^3)$, limiting its applicability to modern large-scale models.
To overcome these limitations, we propose the Matrix Nuclear-Norm as a scalable alternative. By leveraging the $L_{1,2}$ norm as a convex approximation of matrix rank, the Matrix Nuclear-Norm reduces computational complexity to $O(n^2)$. This makes it feasible for evaluating embeddings from large-scale LLMs while preserving the insights provided by Matrix Entropy, such as compression efficiency.

\section{Preliminaries}

This section presents the fundamental concepts for model performance evaluation: discriminability, diversity, and nuclear norm.

\subsection{Discriminability Measurement: F-NORM}

Higher discriminability corresponds to lower prediction uncertainty in the response matrix $A$. When $A$ is normalized as a probability matrix (i.e., $\sum_{j=1}^C A_{i,j} = 1,\ \forall i \in [B]$), this uncertainty can be quantified using Shannon Entropy \citep{shannon1948mathematical}:

\begin{equation}
H(A) = -\frac{1}{B} \sum_{i=1}^B \sum_{j=1}^C A_{i,j} \log\left(A_{i,j}\right)
\end{equation}

where $B$ is the number of samples, $C$ the feature dimension, and $A_{i,j}$ the normalized activation value. Lower entropy indicates higher discriminability. 

An alternative measurement is the Frobenius norm:

\begin{equation}
\|A\|_F = \sqrt{\sum_{i=1}^B \sum_{j=1}^C |A_{i, j}|^2}.
\end{equation}

This norm reflects activation intensity, with higher values indicating more concentrated distributions.

\textbf{Theorem 1.} For a row-normalized matrix $A \in \mathbb{R}_+^{B\times C}$ (i.e., $\sum_{j=1}^C A_{i,j} = 1,\ \forall i$), $H(A)$ and $\|A\|_{F}$ are strictly inversely monotonic. 

The norm satisfies dimensional bounds:
\begin{equation}
\sqrt{\frac{B}{C}} \leq \|A\|_F \leq \sqrt{B}
\end{equation}

where the lower bound achieves when $A$ has uniform distributions (maximal uncertainty), and the upper bound when $A$ contains one-hot vectors (minimal uncertainty). The proof is given in Appendix \ref{sec:Theorem1}.

\subsection{Diversity Measurement: Matrix Rank}
In LLMs, diversity reflects the model's ability to utilize its latent representation space effectively. For a given dataset \( \mathcal{D} \), the expected diversity of outputs is defined as:
\begin{equation}
    E_C = \mathbb{E}_{A \sim \mathcal{D}} \big[ C_p(A) \big]
\end{equation}

To approximate \( C_p(A) \), we construct a sparse matrix \( M \in \{0,1\}^{B \times C} \) where each row contains a one-hot vector indicating the argmax position:
\begin{equation}
    M_{i,j} = \begin{cases}
        1, & j = \arg\max_{k} A_{i,k} \\
        0, & \text{otherwise}
    \end{cases}
\end{equation}

The capacity measure then becomes:
\begin{equation}
    C_p(A) = \mathrm{rank}\big( M \odot A \big) \approx \mathrm{rank}(A)
\end{equation}
where \( \odot \) denotes element-wise product.

The maximum value of \( C_p(A) \) is \( \min(B, C) \), where \( C \) is the output representation dimension. Maximizing \( C_p(A) \) ensures effective utilization of the representation space, promoting robustness through reduced redundancy.

\subsection{Nuclear Norm}

The nuclear norm is an important measure related to diversity and discriminability.

\textbf{Theorem 2.} When \( \|A\| \leq 1 \) (where \( \|A\| \) is the spectral norm), the convex envelope of \( \operatorname{rank}(A) \) is the nuclear norm \( \|A\|_{\star} \). The theorem is proved in \citet{fazel2002matrix}.

For a matrix \( A \in \mathbb{R}^{B \times C} \) with \( \|A\|_{F} \leq \sqrt{B} \), let \( D = \min(B, C) \). The relationships between the nuclear norm and Frobenius norm are:

\begin{equation}
\|A\|_{F} \leq \|A\|_{\star} \leq \sqrt{D} \cdot \|A\|_{F}.
\end{equation}

Therefore, maximizing \( \|A\|_{\star} \) encourages higher rank, which implies high diversity and discriminability. The upper bound of \( \|A\|_{\star} \) is further bounded by:

\begin{equation}
\|A\|_{\star} \leq \sqrt{D \cdot B}.
\end{equation} 

\section{Methodology}

\subsection{Motivation}

Evaluating large language models (LLMs) requires metrics that not only capture model performance but also efficiently handle computational demands. Our initial exploration into Matrix Entropy highlighted its potential as a promising metric for assessing model capabilities, particularly in the realm of information compression. However, its practical application is severely limited by high computational complexity, which escalates with model size, leading to inefficiencies in evaluation.
To overcome these challenges, we propose the Matrix Nuclear-Norm as an alternative, inspired by its relationship with matrix rank—a key component of Matrix Entropy. 
This connection is well-documented in literature, such as \citet{huang2018low} where the nuclear norm effectively approximates matrix rank, thus offering a pathway to mitigate the computational intensity of Matrix Entropy.
Our experiments demonstrate that the Matrix Nuclear-Norm not only reduces computational complexity but also preserves the evaluative strengths of Matrix Entropy. By utilizing the \( L_{1,2}\text{-norm} \) to approximate the nuclear norm, we achieve substantial efficiency gains, ensuring scalability and robustness in LLM evaluation.
Therefore, the Matrix Nuclear-Norm serves as a viable surrogate for Matrix Entropy, providing a comprehensive framework for assessing information compression in large-scale models. This approach allows us to evaluate LLMs more effectively, addressing both theoretical and practical challenges in model assessment.

\subsection{Matrix Nuclear-Norm}

For a matrix \( A \in \mathbb{R}^{B \times C} \), computing its exact nuclear norm via Singular Value Decomposition (SVD) requires \( O(\min(B^2C, BC^2)) \) time, which is equivalent to \( O(n^3) \) with \( n = \max(B, C) \). While feasible for small matrices, this becomes computationally prohibitive for large-scale models. Additionally, numerical instability may arise in SVD computations for ill-conditioned matrices.

\textbf{Sparsity Prior}: When \( A \) exhibits column-wise sparsity (i.e., non-zero activations concentrate in a subset of columns), we can approximate its singular values by leveraging column norms. Let \( \|A\|_F \) denote the Frobenius norm, bounded by \( \|A\|_F \leq \sqrt{\min(B, C)} \cdot \sigma_{\max}(A) \), where \( \sigma_{\max}(A) \) is the largest singular value.

\textbf{Theorem 3.} (Column-Norm Approximation) \\
If \( A \) has rapidly decaying column norms \( \{\|A_{:,j}\|_2\}_{j=1}^C \), the \( j \)-th largest singular value \( \sigma_j(A) \) can be approximated by the \( j \)-th largest column norm:
\begin{equation}
\sigma_j(A) \approx \mathrm{Sort}\left(\left\{\|A_{:,j}\|_2\right\}_{j=1}^C\right)_{[j]}, \quad j \in \{1, \dots, r\},
\end{equation}
where \( r = \mathrm{rank}(A) \). The proof is given in Sect. \ref{sec:Theorem 3} (Supplementary Materials). The nuclear norm is then approximated as:
\begin{equation}
\|\hat{A}\|_{\star} \approx \sum_{j=1}^{D} \mathrm{Sort}\left(\left\{\|A_{:,j}\|_2\right\}_{j=1}^C\right)_{[j]},
\end{equation}
where \( D \leq r \) is a hyperparameter controlling approximation precision, and \( \widetilde{A} \) denotes the column-sparse approximation of \( A \).

\textbf{Remark}: This approximation holds under the assumption that off-diagonal correlations between columns are negligible (i.e., \( A^\top A \approx \mathrm{diag}(\|A_{:,1}\|_2^2, \dots, \|A_{:,C}\|_2^2) \)). For correlated columns, a diagonal correction term may be required.

This approach indicates that the primary components of the \(L_{1,2}\)-norm can effectively approximate the nuclear norm when \(\|A\|_{F}\) is close to \(\sqrt{B}\), while other components can be considered noise. Compared to traditional SVD-based methods (e.g., \citet{guo2015efficient}), this approach reduces computational complexity from \(O(n^3)\) to \(O(n^2)\) and avoids convergence issues by using only standard floating-point operations. The complete algorithm is detailed in Algorithm \ref{tab:algorithm}.

\textbf{Definition of Matrix Nuclear-Norm.} The approach can ultimately be expressed as:
\begin{equation}
\text{Matrix Nuclear-Norm}(\mathbf{X}) = \frac{\sum_{i=1}^{D} \left( \sqrt{\sum_{j=1}^{m} X_{i,j}^2} \right)}{L_{\text{input}}}
\end{equation}

Here, \(L_{\text{input}}\) denotes the length of the input sequence, ensuring comparability through normalization. Our observations indicate that Matrix Nuclear-Norm values increase with longer sequences; further details can be found in Section \ref{sec:Analysis of Length Dynamics}.

\renewcommand{\arraystretch}{2} 

\begin{algorithm}[!h]
    \fontsize{10}{15}\selectfont 
    \renewcommand{\algorithmicrequire}{\textbf{Require:}}
    \caption{Algorithm of Matrix Nuclear-Norm}
    \begin{algorithmic}[1] 
        \REQUIRE Sentence representations $\mathcal{S}=\left\{X_{i}\right\}_{i=1}^{m}$, where \(X_{i}\in \mathbb{R}^{d\times 1}\), \(d\) is the hidden dimension, and \(L_{\text{input}}\) is the sentence length.
        \STATE \(\mu = \frac{1}{m} \sum_{i=1}^{m} X_i\) \textsf{\textcolor{lightgray}{\hfill // Mean embedding}}
        \STATE \(\mathbf{X}_{\text{norm}} = \frac{\mathbf{X} - \mu}{\|\mathbf{X} - \mu\|_{2, \text{row}}}\) \textsf{\textcolor{lightgray}{\hfill // Normalize matrix}}
        \STATE \(\text{L2}(\mathbf{X}_{\text{norm}}) = \sqrt{\sum_{i=1}^{m} \mathbf{X}_{i,j}^2}\) \textsf{\textcolor{lightgray}{\hfill // Column \( L_{2}\text{-norm} \)}}
        \STATE \(\Sigma_D = \{ \sigma_1, \sigma_2, \dots, \sigma_D \}\)  \textsf{\textcolor{lightgray}{\hfill // Top \( D \) norms}}
        \STATE \(\text{Matrix Nuclear-Norm}(\mathbf{X}) = \frac{\sum_{i=1}^{D} \left( \sqrt{\sum_{j=1}^{m} \mathbf{X}_{j,i}^2} \right)}{L_{\text{input}}}\)  \textsf{\textcolor{lightgray}{}}
        
        \STATE \textbf{return} \text{Matrix Nuclear-Norm} 
    \end{algorithmic}
    \label{tab:algorithm}
\end{algorithm}

\section{Experiments of Large Language Models}
The models and datasets used in this paper are thoroughly introduced in ~\ref{sec:model and dataset}.

\subsection{Baselines}

\textbf{Cross-Entropy Loss.} Cross-entropy is a key metric for evaluating LLMs by measuring the divergence between predicted and true probability distributions. The formula is given as \citep{wei2024large}:

\begin{equation}
\mathcal{L}_{\text{CE}} = -\frac{1}{T} \sum_{i=1}^{T} \log P(u_i \mid u_{<i}; \Theta)
\end{equation}
where \( u_i \) is the target token at position \( i \), \( P(u_i \mid u_{<i}; \Theta) \) is the conditional probability predicted by the model, and \( T \) is the sequence length. Lower values indicate better prediction accuracy. We compare this baseline with the Matrix Nuclear Norm metric, using the same datasets and models from \citep{kaplan2020scaling}.

\textbf{Perplexity.} Perplexity measures how well a language model predicts a sequence of words. For a text sequence \( \mathbf{U} = \{u_1, \ldots, u_T\} \), it is defined as \citep{neubig2017neural,wei2024large}:

\begin{equation}
\text{PPL}(\mathbf{U}) = \exp\left( -\frac{1}{T} \sum_{i=1}^{T} \log P(u_i \mid u_{<i}; \Theta) \right)
\end{equation}

Lower perplexity indicates better performance, showing that fewer attempts are needed to predict the next token.

\textbf{Matrix Entropy of a Dataset.} For a dataset \( \mathcal{D} = \{ \mathbf{S}_i \}_{i=1}^n \), where \( \mathbf{S}_i \in \mathbb{R}^{d \times d} \) represents sentence embedding covariance matrices, the normalized matrix entropy is defined as \citep{wei2024large}:

\begin{equation}
H(\mathcal{D}) = \frac{1}{n \log d} \sum_{i=1}^{n} H\left( \frac{\sigma(\mathbf{S}_i)}{\|\sigma(\mathbf{S}_i)\|_1} \right)
\end{equation}
where \( \sigma(\mathbf{S}_i) \) denotes the singular values of matrix \( \mathbf{S}_i \), and \( H(\cdot) \) is the Shannon entropy computed over the normalized singular value distribution.

\subsubsection{Language Models}

In our experiments, we selected a range of widely used transformer-based LLMs. Notably, we included Cerebras-GPT \citep{gao2020pile}, a pre-trained model well-suited for studying scaling laws. The selection of Cerebras-GPT is particularly advantageous due to its diverse model sizes, which span from 111 million to 13 billion parameters. This diversity allows for a comprehensive analysis of pre-trained language models across varying scales. Additionally, we utilized various scaled versions of the Pythia model \citep{biderman2023pythia}, ranging from 14 million to 12 billion parameters, to further examine performance variations as model scale changes, thus validating the effectiveness of the proposed Matrix Nuclear-Norm metric.

We conducted Matrix Nuclear-Norm calculations and comparative analyses on inference responses from these models using two benchmark datasets: AlpacaEval and ChatBot Arena. The specific models included in our study are the DeepSeek series \citep{deepseek-coder} (1.3B, 6.7B, 7B), the Llama3 series \citep{dubey2024llama} (8B, 70B), the QWEN 2 series \citep{yang2024qwen2} (0.5B, 1.5B, 7B, 72B), and the Vicuna series \citep{chiang2023vicuna} (7B, 13B, 33B). We also evaluated models of the same scale, specifically Gemma-7B \citep{team2024gemma} and Mistral-7B \citep{jiang2023mistral}. The inclusion of these diverse models enriches our research perspective and facilitates an in-depth exploration of the inference performance and scaling laws of LLMs across different parameter sizes.

\subsection{Matrix Nuclear-Norm Observation}

\subsubsection{Comparing Computational Time}

\begin{figure}[t]
    \centering
    \includegraphics[width=0.5\textwidth]{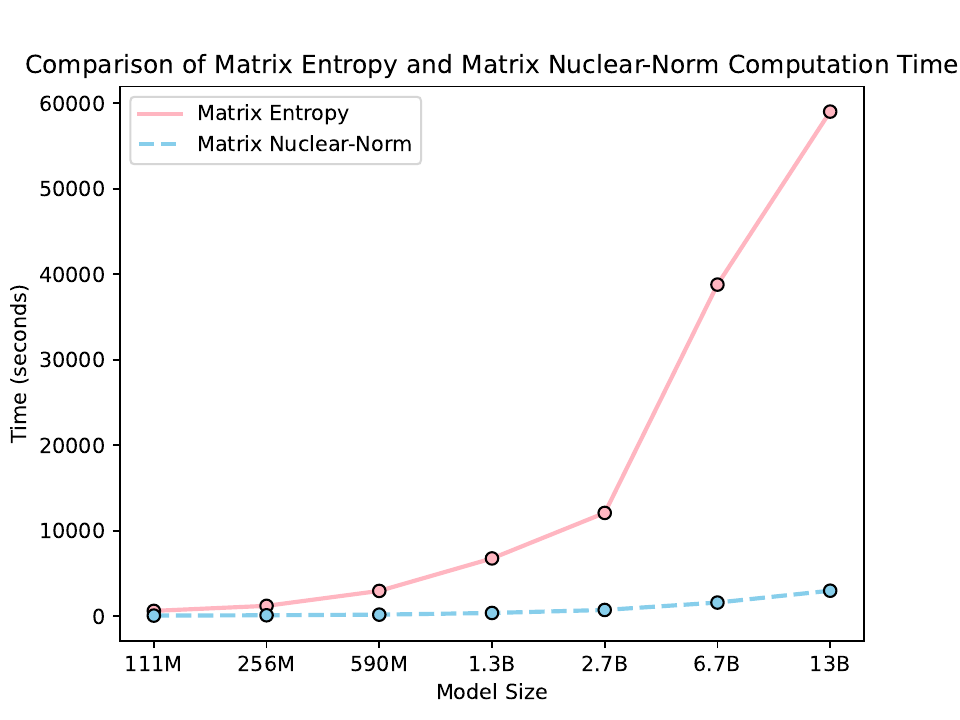} 
    \caption{Cerebras-GPT: Time comparison}
    \label{fig:time-gpt-fig}
    \vspace{-5pt}
\end{figure} 
To evaluate the computational efficiency of Matrix Nuclear-Norm in comparison to Matrix Entropy for LLMs, we conducted experiments across various model sizes using multiple benchmark datasets. The results, summarized in Table \ref{tab:time-gpt}, demonstrate a clear advantage of Matrix Nuclear-Norm in terms of computation time, particularly for larger models.

As model sizes increased, Matrix Entropy's computation time rose dramatically, reaching approximately 16.3 hours for the 13B model . In contrast, Matrix Nuclear-Norm only required about 0.82 hours for the same model, representing nearly a 20-fold reduction in computation time. This trend was consistent across all model sizes, with Matrix Nuclear-Norm consistently proving to be much faster (as illustrated in Figure \ref{fig:time-gpt-fig}). For example, the 111M model showed that Matrix Nuclear-Norm was 8.58 times quicker than Matrix Entropy.

The significant efficiency gain is due to the lower complexity of Matrix Nuclear-Norm, $O(m \cdot n + n \log n)$, versus Matrix Entropy’s $O(n^3)$, where $m$ is the embedding dimension (columns). This makes it an efficient metric for LLM evaluation, especially for large-scale models.

In summary, Matrix Nuclear-Norm achieves comparable evaluation accuracy to Matrix Entropy but with vastly superior computational efficiency, making it a practical and scalable choice for assessing LLMs.

\renewcommand{\arraystretch}{1}
\begin{table}[ht]
\centering
\resizebox{\linewidth}{!}{ 
\begin{tabular}{cccc}
\hline
\multirow{1}{*}{\textbf{Model Size}} & \multirow{1}{*}{\textbf{ME Time(s)}} & \multirow{1}{*}{\textbf{MNN Time(s)}} & \multirow{1}{*}{\textbf{Ratio}} \\
\midrule
{111M}& {623.5}& {72.7}& {8.6} \\ 
{256M}& {1213.1}& {110.8}& {10.9}\\
{590M}& {2959.7}& {184.8}& {16.0}\\
{1.3B}& {6760.2}& {379.0}& {17.8}\\
{2.7B}& {12083.7}& {732.6}& {16.5}\\
{6.7B}& {38791.2}& {1598.4}& {\textbf{24.3}}\\
{13B} & {59028.4}& {2984.2}& {19.8}\\ \bottomrule
\end{tabular}
}
\caption{Cerebras-GPT: Time Comparison between Matrix Entropy (ME) and Matrix Nuclear-Norm (MNN)}
\label{tab:time-gpt}
\end{table}

\subsubsection{Scaling Law of Matrix Nuclear-Norm}
\label{sec:Scaling Law of Matrix Nuclear-Norm}
\begin{figure}[t]
    \centering
    \begin{subfigure}[b]{0.42\textwidth}
        \includegraphics[width=\textwidth]{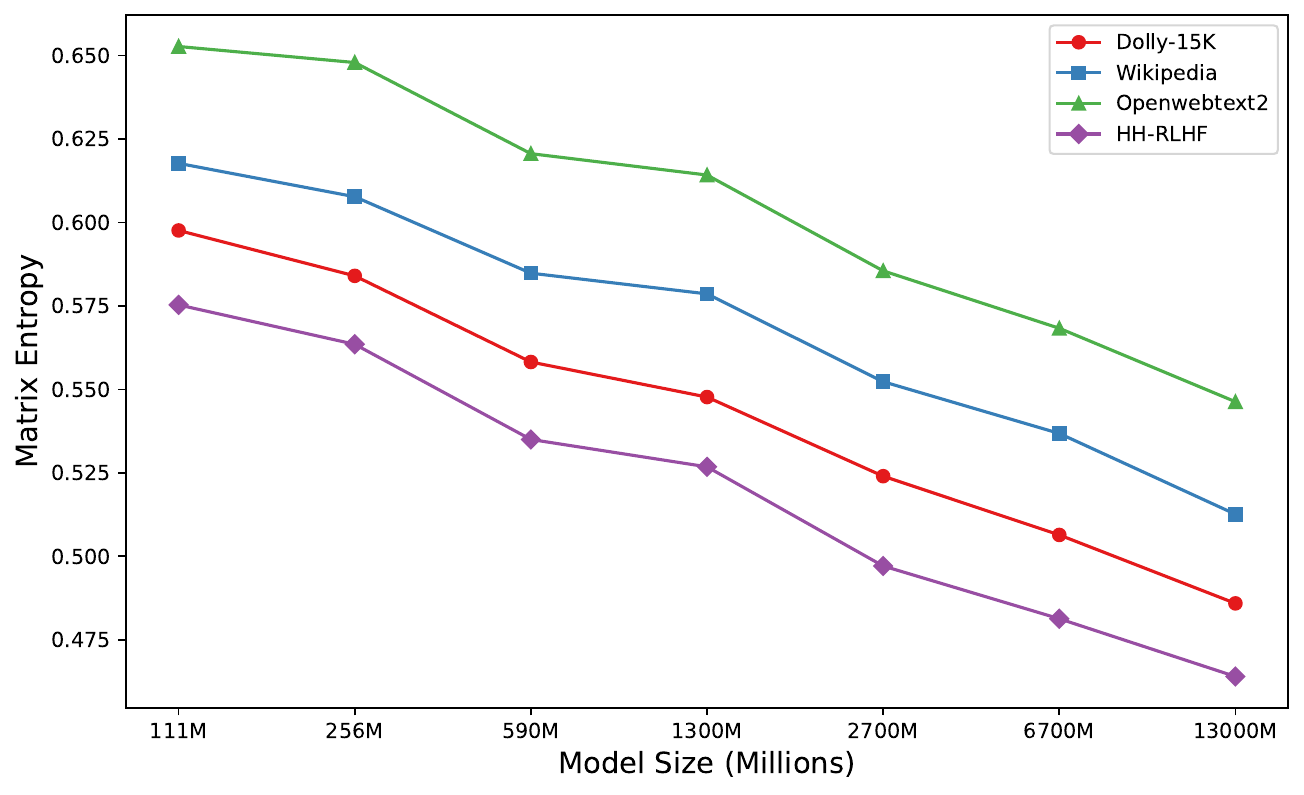} 
        \caption{Matrix Entropy}
        \label{fig:matrix-entropy-gpt-fig}
    \end{subfigure}
    \begin{subfigure}[b]{0.42\textwidth}
        \includegraphics[width=\textwidth]{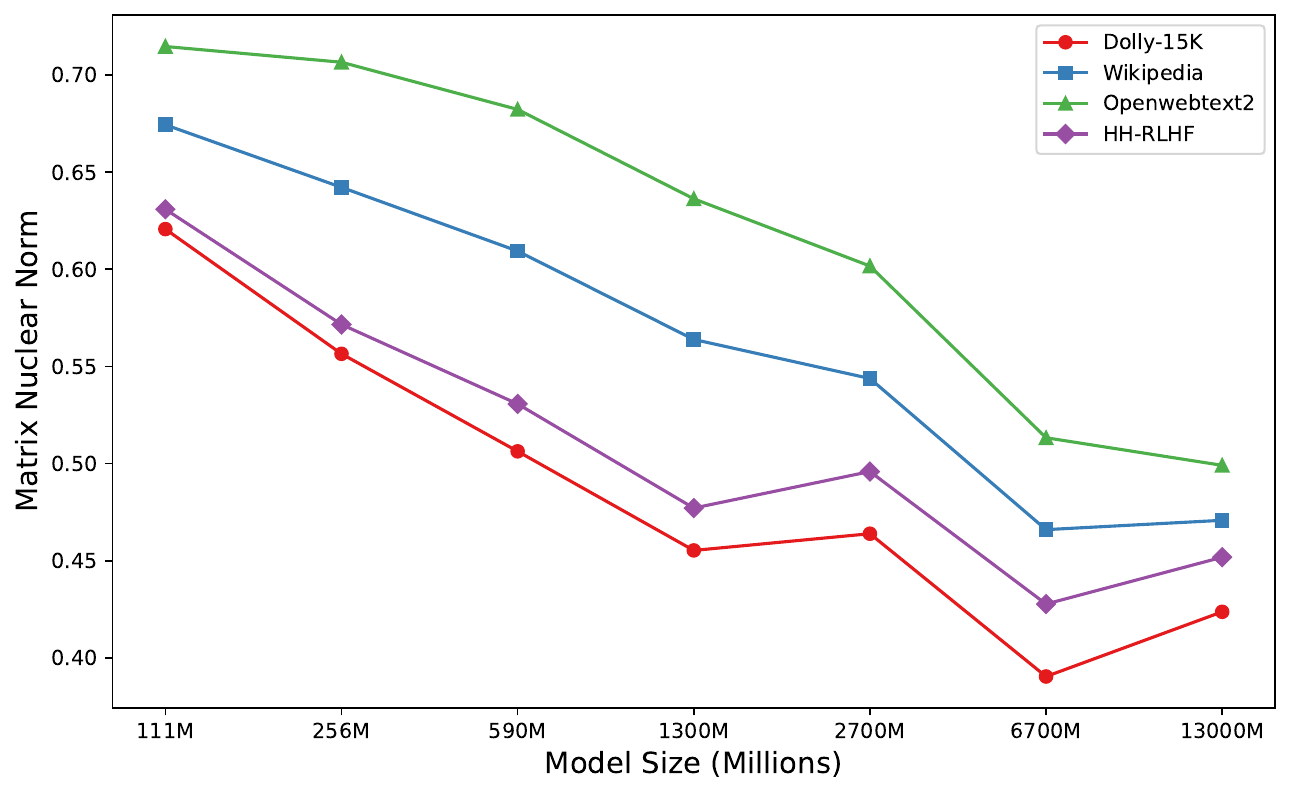} 
        \caption{Matrix Nuclear-Norm}
        \label{fig:Matrix Nuclear-Norm-gpt-fig}
    \end{subfigure}
    \caption{Comparison of Matrix Nuclear-Norm, matrix entropy when model scales up.} 
\end{figure}

To affirm Matrix Nuclear-Norm's efficacy as an evaluative metric, we evaluated Cerebras-GPT models on four datasets including dolly-15k, Wikipedia, openwebtext2, and hh-rlhf comparing Matrix Nuclear-Norm, matrix entropy, perplexity, and loss. As shown in Table \ref{tab:four-dataset-gpt}, Matrix Nuclear-Norm decreases consistently with model size, indicating better data compression and processing in larger models. This trend (Figure \ref{fig:Matrix Nuclear-Norm-gpt-fig}) validates Matrix Nuclear-Norm's utility across datasets. Notably, anomalies at the 2.7B and 13B highlight areas needing further exploration.

\subsubsection{Relationship of Benchmark Indicators}

Findings indicate the efficacy of the Matrix Nuclear-Norm as a metric for evaluating LLM, as shown in Table \ref{tab:Three-dataset-gpt} (Appendix), there is an overall downward trend in Matrix Nuclear-Norm values with increasing model sizes, signifying enhanced compression efficiency. However, notable anomalies at the 2.7B and 13B checkpoints suggest that these specific model sizes warrant closer examination. Despite these discrepancies, the Matrix Nuclear-Norm consistently demonstrates superior computational efficiency and accuracy compared to traditional metrics, highlighting its promising applicability for future model evaluations.

\subsection{Language Investigation}

\subsubsection{Sentence Operation Experiments}

\begin{figure}[t]
    \centering
    \includegraphics[width=0.4\textwidth]{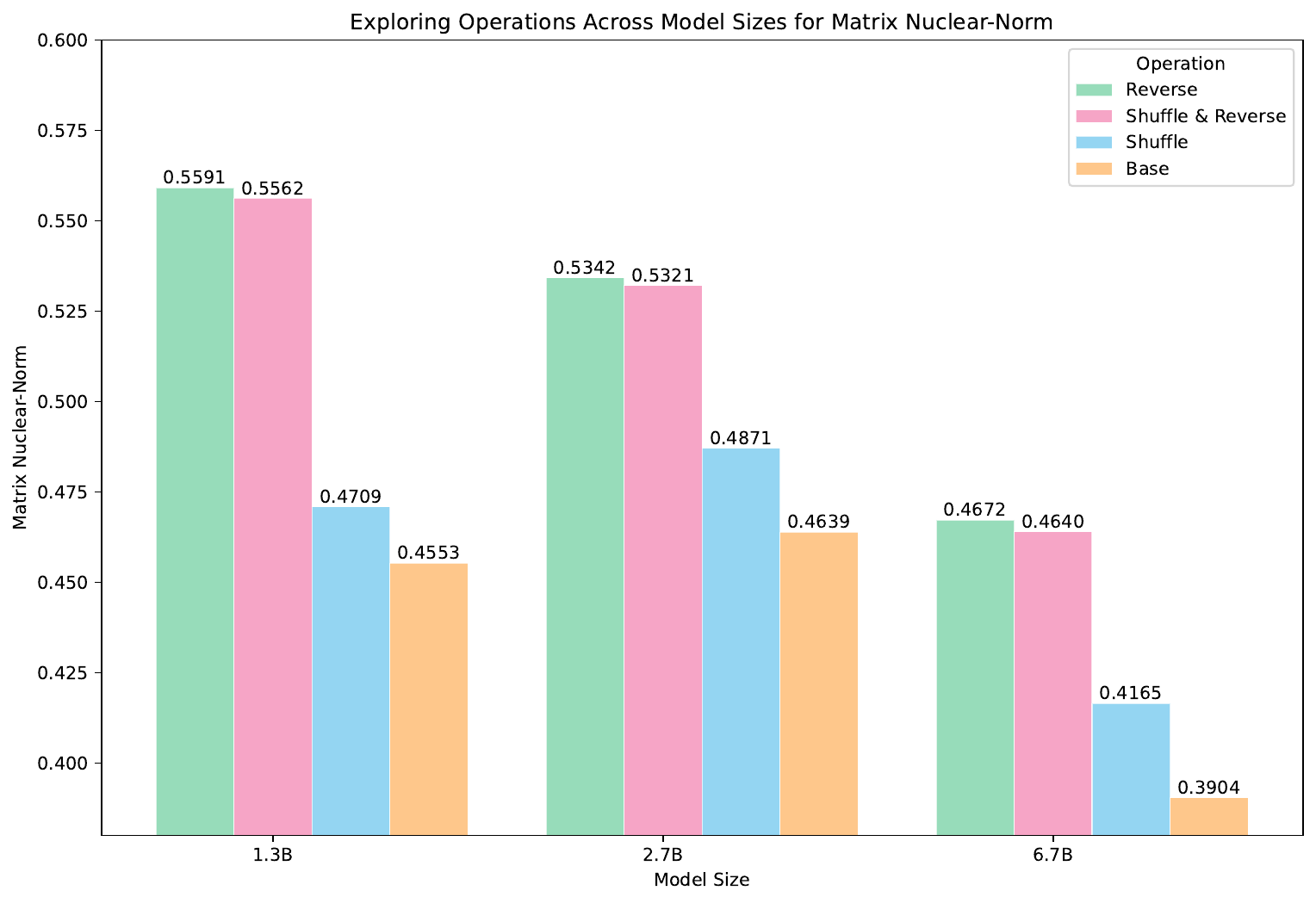} 
    \caption{Results of sentence operation. Shuffling and reversing disrupt the text structure and diminish the informational content, leading to an increase in Matrix Nuclear-Norm.}
    \label{fig:sentence operation}
\end{figure}

Figure \ref{fig:sentence operation} shows sentence manipulations impact Matrix Nuclear-Norm values. These values decrease with model size, in line with established scaling laws similar to those governing matrix entropy and perplexity (PPL). As models grow larger, they can capture data patterns more efficiently, reducing redundant information representation, which directly lowers the nuclear norm.

The ranking Reverse $>$ Shuffle \& Reverse $>$ Shuffle $>$ Base reflects input disruption. Reverse flips the sentence, introducing maximum disorder and causing a large norm increase. Shuffle only partially rearranges elements, leading to a smaller rise. The unaltered Base condition enables optimal compression.

Notably, the 2.7B model has slightly higher Shuffle and Base values than the 1.3B model, yet this doesn't challenge the conclusion that larger models compress better. The norm increases with text length because longer texts carry more information, increasing entropy and computational complexity. More data means more potential redundancy for the model to process, driving up the norm value. These results clarify model behavior in relation to size, input structure, and length.

\subsubsection{Analysis of Length Dynamics}
\label{sec:Analysis of Length Dynamics}
\begin{figure}[t]
    \centering
    \includegraphics[width=0.4\textwidth]{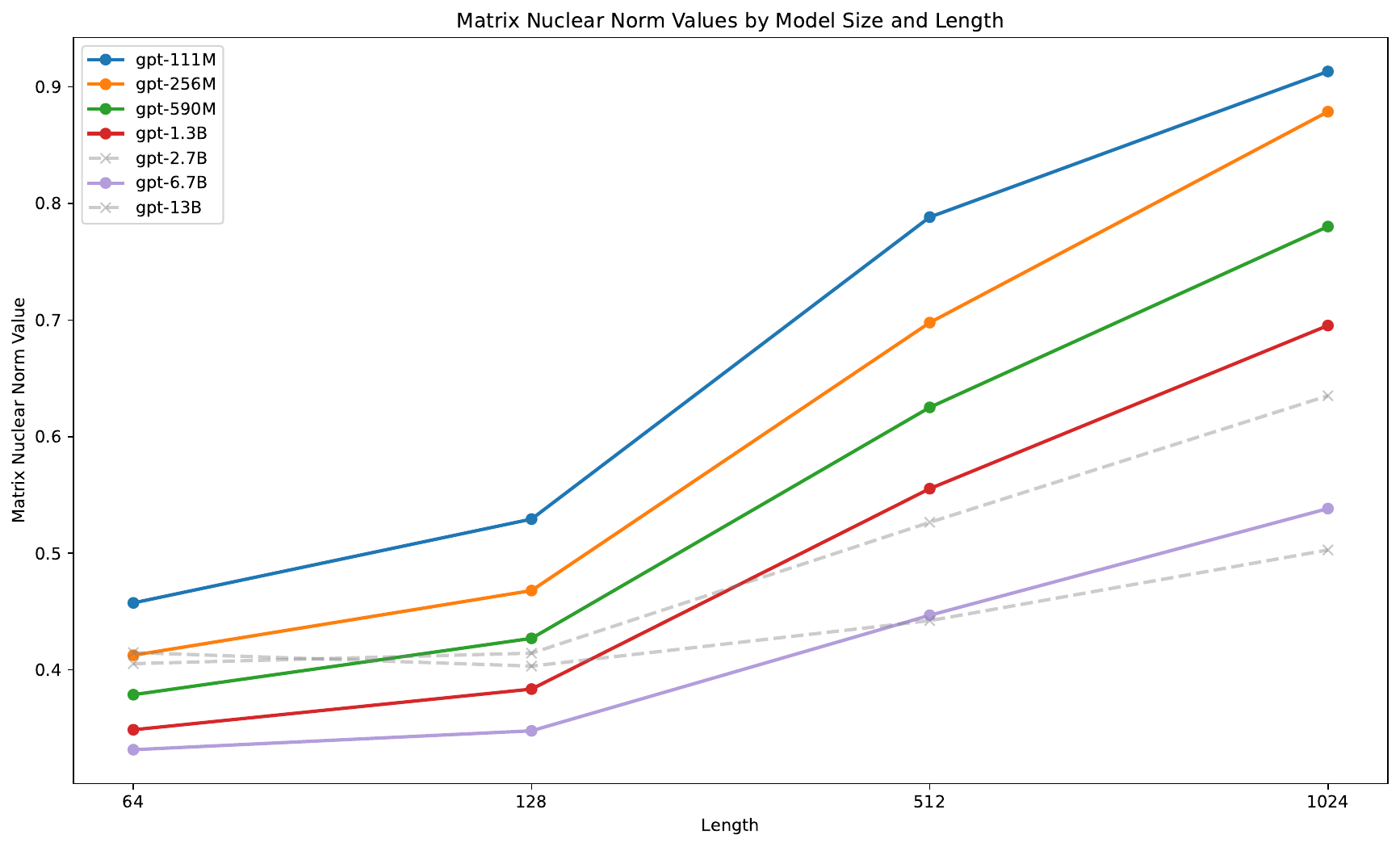} 
    \caption{The Matrix Nuclear-Norm values increase consistently with longer text input lengths, reflecting the model's ability to capture more information.}
    \label{fig:Length Dynamics-fig}
    \vspace{-5pt}
\end{figure}

The analysis reveals that Matrix Nuclear-Norm generally increase as input length rises, aligning with our expectations (see Figure \ref{fig:Length Dynamics-fig}). Longer inputs necessitate that the model manage and compress more information, which naturally leads to higher Matrix Nuclear-Norm. Most models exhibit this trend, indicating effective handling of the increased information load.

However, the Cerebras-GPT-2.7B and Cerebras-GPT-13B models display anomalies in their Matrix Nuclear-Norm values at 64 and 128 tokens, where the value at 128 tokens is lower than that at 64 tokens. This discrepancy may be attributed to these models employing different information compression mechanisms or optimization strategies tailored to specific input lengths, allowing for more effective compression at those lengths.

Overall, aside from a few outliers, the results largely conform to expectations, demonstrating that Matrix Nuclear-Norm values increase with input length, reflecting the greater volume and complexity of information that models must handle.To address the observed trend of rising Matrix Nuclear-Norm values with longer sentences, we incorporated a normalization step in our methodology via dividing the Matrix Nuclear-Norm values by the sentence length. This adjustment helps mitigate any biases introduced by models that tend to generate longer sentences during inference.


\subsubsection{Analysis of Prompt Learning}
The experimental results (shown in Table \ref{tab:result-prompt}) indicate that we performed inference on different sizes of Cerebras-GPT models using three carefully selected prompts (shown in Table \ref{tab:three-prompt}) and calculated the Matrix Nuclear-Norm values of their responses. As the model size increased, the Matrix Nuclear-Norm values gradually decreased, demonstrating that larger models possess greater information compression capabilities. The prompts significantly influenced Matrix Nuclear-Norm, with variations reflecting the models' responses to prompt complexity. Specifically, Cerebras-GPT-1.3B showed a notable decrease in Matrix Nuclear-Norm after the input prompts, indicating its sensitivity to them, while Cerebras-GPT-2.7B exhibited smaller changes. In contrast, Cerebras-GPT-6.7B displayed minimal variation across all prompts, suggesting stable performance regardless of prompt detail. Overall, more detailed prompts resulted in larger information volumes in the model's responses, leading to corresponding changes in Matrix Nuclear-Norm values.

\renewcommand{\arraystretch}{0.9}
\begin{table}[ht]
\caption{Results of prompt learning without Prompt and with (Prompt 1, 2, 3) the use of prompts. Incorporating prompts as prefixes before the QA pairs enhances the models' ability to achieve better compression.}
\label{tab:result-prompt}
\noindent 
\resizebox{\linewidth}{!}{%
    \begin{tabular}{@{}ccccccc@{}}
    \toprule
    & \multicolumn{6}{c}{ADDING PROMPT TO QA PAIRS} \\ \cmidrule(l){2-7} 
    \multirow{-2}{*}{MODELS} & EMPTY PROMPT & PROMPT 1 & PROMPT 2 & PROMPT 3 & AVERAGE & \(\Delta x\) \\ \midrule
    Cerebras-GPT-1.3B & 0.150955 & 0.147577 & 0.140511 & 0.141358 & 0.14453 & \textcolor{red}{\textbf{↓0.006425}} \\ \midrule
    Cerebras-GPT-2.7B & 0.150130 & 0.151522 & 0.142834 & 0.151842 & 0.14844 & \textcolor{red}{\textbf{↓0.001690}} \\ \midrule
    Cerebras-GPT-6.7B & 0.132042 & 0.128346 & 0.124094 & 0.133211 & 0.12923 & \textcolor{red}{\textbf{↓0.002812}} \\ \bottomrule
    \end{tabular}%
}
\vspace{-10pt}
\end{table}

\section{Evaluating and Ranking LLMs}
\subsection{Inference-Based Model Assessment}
In this section, we evaluated model inference across the AlpacaEval and Chatbot Arena benchmarks using the Matrix Nuclear-Norm metric prior to the final MLP classification head. The analysis revealed that Matrix Nuclear-Norm reliably ranks model performance, with lower values indicating enhanced information processing efficiency, particularly as model size scales up.

For instance, the Llama-3 70B model demonstrated superior compression capabilities compared to its 8B counterpart, as reflected by significantly lower Matrix Nuclear-Norm values across both benchmarks (see Table \ref{tab:Matrix Nuclear-Norm of Llama-3}). A similar trend was observed in the Vicuna family, where Matrix Nuclear-Norm values consistently decreased from 0.4623 for the 7B model to 0.3643 for the 33B model on the AlpacaEval dataset, indicating progressive improvements in information handling (see Table \ref{tab:Nuclear-Norms-Vicuna-DeepSeek}). Additionally, the DeepSeek models exhibited a consistent decrease in Matrix Nuclear-Norm values as model size increased, further demonstrating the metric’s validity.

Overall, these results substantiate Matrix Nuclear-Norm as a robust and reliable tool for evaluating and ranking LLMs, demonstrating its capacity to capture critical aspects of model performance across diverse benchmarks.

\begin{table}[ht]
\centering

\resizebox{0.48\textwidth}{!}{ 
\begin{tabular}{ccccc}
\toprule
\textbf{Model} & \textbf{Data}  & \textbf{1.3B} & \textbf{6.7B} & \textbf{7B} \\ 
\midrule
\multirow{2}{*}{DeepSeek} & Alpaca  & 0.4882 & 0.3472 & 0.3352 \\
& Arena & 0.5754 & 0.4175 & 0.4357 \\ 
\multirow{2}{*}{Vicuna} & Alpaca & 0.4623 & 0.4159 & 0.3643 \\
& Arena & 0.4824 & 0.4311 & 0.3734 \\ 
\bottomrule
\end{tabular}
}

\caption{Matrix Nuclear-Norms in Vicuna and DeepSeek Responses}
\vspace{-10pt}
\label{tab:Nuclear-Norms-Vicuna-DeepSeek}
\end{table}


\subsection{Matrix Nuclear-Norm for Model Ranking}
In this experimental section, we utilized Matrix Nuclear-Norm to evaluate the responses of LLMs, focusing on 7B and 70B variants. Notably, lower Matrix Nuclear-Norm values indicate more efficient information compression, serving as a robust indicator of model performance.

Among the 7B models, DeepSeek-7B exhibited the most efficient information processing with the lowest average Matrix Nuclear-Norm score of 0.3855 across Alpaca and Arena datasets (see Table \ref{tab:Nuclear-Norms-Vicuna-DeepSeek}). Gemma-7B followed closely with an average score of 0.3879, whereas QWEN 2-7B demonstrated less efficient compression with an average score of 0.5870. In contrast, the 70B models showed varied performance, with Llama 2-70B achieving the best average score of 0.3974, slightly outperforming Llama 3-70B (0.4951) and QWEN models, which scored around 0.5.

Interestingly, certain 7B models, like DeepSeek-7B and Gemma-7B, outperformed larger 70B models, underscoring that model efficiency is not solely determined by size. These results highlight that factors such as architecture, training methodology, and data complexity play crucial roles in information processing capabilities beyond scale.

\begin{table}[ht]
\centering
\resizebox{\linewidth}{!}{%
\begin{tabular}{lcccc}
\cmidrule(lr){1-4} 
\multirow{2}{*}{\centering\vspace*{-2mm}\hspace*{5mm}\textbf{MODEL}} 
 & \multicolumn{3}{c}{\textbf{Matrix Nuclear-Norm}} \\
\cmidrule(lr){2-4} 
& \textbf{Alpaca} & \textbf{Arena-Hard} & \textbf{Avg Score}  \\ 
\cmidrule(lr){1-4} 
QWEN 2-7B & 0.5989 & 0.5751 & 0.5870  \\ 
Mistral-7B & 0.4980 & 0.5126 & 0.5053   \\
QWEN 1.5-7B & 0.4866 & 0.5165 & 0.5016   \\
LLaMA 2-7B & 0.4648 & 0.5038 & 0.4843   \\
Vicuna-7B & 0.4623 & 0.4824 & 0.4724   \\
Gemma-7B & 0.3759 & 0.3998 & 0.3879   \\
\textbf{DeepSeek-7B} & 0.3352 & 0.4357 & \textbf{0.3855}   \\

\cmidrule(lr){1-4} 

QWEN 1.5-72B & 0.5291 & 0.5065 & 0.5178  \\ 
QWEN 2-72B & 0.5261 & 0.4689 & 0.4975  \\
Llama 3-70B & 0.4935 & 0.4967 & 0.4951  \\
\textbf{Llama 2-70B} & 0.3862 & 0.4086 & \textbf{0.3974}  \\ 
\cmidrule(lr){1-4} 
\end{tabular}%
}
\caption{Descending Competence Rankings via Matrix Nuclear Norm: Small and Large LMs}
\label{tab:sort}
\end{table}

To validate the design rationale and robustness of the Matrix Nuclear-Norm, we conducted a series of ablation studies. Due to space constraints, detailed results are provided in \ref{sec:ablation-study} (appendix) to maintain brevity in the main text. These experiments included evaluations across different model families, such as Cerebras-GPT and Pythia, as well as comparisons of various data sampling strategies.The results demonstrate that the Matrix Nuclear-Norm consistently performs well across different model scales and sampling variations. This not only confirms its applicability across diverse models but also verifies its stability and reliability in handling large-scale datasets. We also provide an ablation Cerebras-GPT:  study in the appendix, further proving the method’s efficiency and accuracy in evaluating LLMs.

\section{Conclusion}
In conclusion, Matrix Nuclear-Norm stands out as a promising evaluation metric for LLMs, offering significant advantages in assessing information compression and redundancy elimination. Its key strengths include remarkable computational efficiency, greatly exceeding that of existing metrics like matrix entropy, along with exceptional stability across diverse datasets. Matrix Nuclear-Norm’s responsiveness to model performance under varying inputs emphasizes its ability to gauge not only performance but also the intricate adaptability of models. This metric marks a significant advancement in NLP, establishing a clear and effective framework for future research and development in the evaluation and optimization of language models.

\section{Limitations}

Although Matrix Nuclear-Norm (MNN) performs well in evaluating LLM performance, it has three main limitations. First, as MNN computation relies on hidden states, the results are sensitive to model architecture and training processes. This may cause performance inconsistencies across different model designs or training settings (particularly between Cerebras-GPT-1.3B and Cerebras-GPT-2.7B), potentially limiting broader applicability. Second, while MNN offers computational advantages over traditional methods, it may still face resource challenges when evaluating extremely large models, requiring further optimization for scalability.

Third, our current implementation uses MNN primarily as an evaluation metric rather than a training objective. However, we recognize its potential for analyzing information compression dynamics during training, which could provide valuable insights into model optimization. Future work should explore this direction while addressing the method's sensitivity to architectural variations.

Notably, despite observed anomalies in specific configurations, MNN demonstrates consistent computational efficiency and accuracy across various model sizes and data sampling strategies. We will enhance our discussion of these performance variations to better clarify the method's robustness boundaries and operational constraints. These limitations highlight the need for continued research into architecture-agnostic evaluation frameworks and optimized computation strategies as language models scale.
\section{Ethics Statement}
Our study adheres to strict ethical guidelines by utilizing only publicly available and open-source datasets. We ensured that all datasets used, such as dolly-15k, hh-rlhf, OpenBookQA, Winogrande, PIQA, AlpacaEval, and Chatbot Arena, are free from harmful, biased, or sensitive content. Additionally, careful curation was conducted to avoid toxic, inappropriate, or ethically problematic data, thereby ensuring the integrity and safety of our research. This commitment reflects our dedication to responsible AI research and the broader implications of using such data in language model development.

\section{Reproducibility}
We emphasize the importance of reproducibility in the development and evaluation of our newly proposed metric, Matrix Nuclear-Norm. To facilitate reproducibility, we provide detailed information regarding our data processing and parameter settings:

\textbf{Data Processing and Parameter Settings:} We outline the preprocessing steps applied to each dataset, ensuring that other researchers can accurately replicate our methodology. All hyperparameters and configuration settings used during the experiments are specified in the code, offering clarity on the experimental conditions.

\textbf{Experimental Procedures:} We detail the specific steps required to evaluate the Matrix Nuclear-Norm, including its application to each dataset and the metrics used for performance assessment.

\textbf{Code Availability:} Our implementation code, evaluation scripts, and pretrained models will be made publicly available upon acceptance of this paper, enabling others to reproduce our experiments and validate our findings.

By adhering to these guidelines, we aim to ensure that our work is accessible and reproducible for future research endeavors.





\bibliography{main}

\begin{thebibliography}{44}
\providecommand{\natexlab}[1]{#1}

\bibitem[{Bai et~al.(2022)Bai, Jones, Ndousse, Askell, Chen, DasSarma, Drain, Fort, Ganguli, Henighan et~al.}]{bai2022training}
Yuntao Bai, Andy Jones, Kamal Ndousse, Amanda Askell, Anna Chen, Nova DasSarma, Dawn Drain, Stanislav Fort, Deep Ganguli, Tom Henighan, et~al. 2022.
\newblock Training a helpful and harmless assistant with reinforcement learning from human feedback.
\newblock \emph{arXiv preprint arXiv:2204.05862}.

\bibitem[{Biderman et~al.(2023)Biderman, Schoelkopf, Anthony, Bradley, O’Brien, Hallahan, Khan, Purohit, Prashanth, Raff et~al.}]{biderman2023pythia}
Stella Biderman, Hailey Schoelkopf, Quentin~Gregory Anthony, Herbie Bradley, Kyle O’Brien, Eric Hallahan, Mohammad~Aflah Khan, Shivanshu Purohit, USVSN~Sai Prashanth, Edward Raff, et~al. 2023.
\newblock Pythia: A suite for analyzing large language models across training and scaling.
\newblock In \emph{International Conference on Machine Learning}, pages 2397--2430. PMLR.

\bibitem[{Bisk et~al.(2020)Bisk, Zellers, Gao, Choi et~al.}]{bisk2020piqa}
Yonatan Bisk, Rowan Zellers, Jianfeng Gao, Yejin Choi, et~al. 2020.
\newblock Piqa: Reasoning about physical commonsense in natural language.
\newblock In \emph{Proceedings of the AAAI conference on artificial intelligence}, volume~34, pages 7432--7439.

\bibitem[{Chen et~al.(2024)Chen, Mao, Li, Jin, Wen, Wei, Wang, Yin, Fan, Liu et~al.}]{chen2024exploring}
Zhikai Chen, Haitao Mao, Hang Li, Wei Jin, Hongzhi Wen, Xiaochi Wei, Shuaiqiang Wang, Dawei Yin, Wenqi Fan, Hui Liu, et~al. 2024.
\newblock Exploring the potential of large language models (llms) in learning on graphs.
\newblock \emph{ACM SIGKDD Explorations Newsletter}, 25(2):42--61.

\bibitem[{Chiang et~al.(2023)Chiang, Li, Lin, Sheng, Wu, Zhang, Zheng, Zhuang, Zhuang, Gonzalez et~al.}]{chiang2023vicuna}
Wei-Lin Chiang, Zhuohan Li, Zi~Lin, Ying Sheng, Zhanghao Wu, Hao Zhang, Lianmin Zheng, Siyuan Zhuang, Yonghao Zhuang, Joseph~E Gonzalez, et~al. 2023.
\newblock Vicuna: An open-source chatbot impressing gpt-4 with 90\%* chatgpt quality.
\newblock \emph{See https://vicuna. lmsys. org (accessed 14 April 2023)}, 2(3):6.

\bibitem[{Chiang et~al.(2024)Chiang, Zheng, Sheng, Angelopoulos, Li, Li, Zhang, Zhu, Jordan, Gonzalez et~al.}]{chiang2403chatbot}
Wei-Lin Chiang, Lianmin Zheng, Ying Sheng, Anastasios~Nikolas Angelopoulos, Tianle Li, Dacheng Li, Hao Zhang, Banghua Zhu, Michael Jordan, Joseph~E Gonzalez, et~al. 2024.
\newblock Chatbot arena: An open platform for evaluating llms by human preference, 2024.
\newblock \emph{URL: https://arxiv. org/abs/2403.04132}.

\bibitem[{Conover et~al.(2023)Conover, Hayes, Mathur, Xie, Wan, Shah, Ghodsi, Wendell, Zaharia, and Xin}]{conover2023free}
Mike Conover, Matt Hayes, Ankit Mathur, Jianwei Xie, Jun Wan, Sam Shah, Ali Ghodsi, Patrick Wendell, Matei Zaharia, and Reynold Xin. 2023.
\newblock Free dolly: Introducing the world’s first truly open instruction-tuned llm.
\newblock \emph{Company Blog of Databricks}.

\bibitem[{Del{\'e}tang et~al.(2023)Del{\'e}tang, Ruoss, Duquenne, Catt, Genewein, Mattern, Grau-Moya, Wenliang, Aitchison, Orseau et~al.}]{deletang2023language}
Gr{\'e}goire Del{\'e}tang, Anian Ruoss, Paul-Ambroise Duquenne, Elliot Catt, Tim Genewein, Christopher Mattern, Jordi Grau-Moya, Li~Kevin Wenliang, Matthew Aitchison, Laurent Orseau, et~al. 2023.
\newblock Language modeling is compression.
\newblock \emph{arXiv preprint arXiv:2309.10668}.

\bibitem[{Dubey et~al.(2024)Dubey, Jauhri, Pandey, Kadian, Al-Dahle, Letman, Mathur, Schelten, Yang, Fan et~al.}]{dubey2024llama}
Abhimanyu Dubey, Abhinav Jauhri, Abhinav Pandey, Abhishek Kadian, Ahmad Al-Dahle, Aiesha Letman, Akhil Mathur, Alan Schelten, Amy Yang, Angela Fan, et~al. 2024.
\newblock The llama 3 herd of models.
\newblock \emph{arXiv preprint arXiv:2407.21783}.

\bibitem[{Dubois et~al.(2024)Dubois, Galambosi, Liang, and Hashimoto}]{dubois2024length}
Yann Dubois, Bal{\'a}zs Galambosi, Percy Liang, and Tatsunori~B Hashimoto. 2024.
\newblock Length-controlled alpacaeval: A simple way to debias automatic evaluators.
\newblock \emph{arXiv preprint arXiv:2404.04475}.

\bibitem[{Fazel(2002)}]{fazel2002matrix}
Maryam Fazel. 2002.
\newblock \emph{Matrix rank minimization with applications}.
\newblock Ph.D. thesis, PhD thesis, Stanford University.

\bibitem[{{Foundation}(2024)}]{wikimedia_downloads}
{Foundation}. 2024.
\newblock {Foundation}.
\newblock {\url{https://dumps.wikimedia.org}}.
\newblock [Online; accessed 2024-09-27].

\bibitem[{Gao et~al.(2020)Gao, Biderman, Black, Golding, Hoppe, Foster, Phang, He, Thite, Nabeshima et~al.}]{gao2020pile}
Leo Gao, Stella Biderman, Sid Black, Laurence Golding, Travis Hoppe, Charles Foster, Jason Phang, Horace He, Anish Thite, Noa Nabeshima, et~al. 2020.
\newblock The pile: An 800gb dataset of diverse text for language modeling.
\newblock \emph{arXiv preprint arXiv:2101.00027}.

\bibitem[{Gemini et~al.(2023)Gemini, Anil, Borgeaud, Wu, Alayrac, Yu, Soricut, Schalkwyk, Dai, Hauth et~al.}]{team2023gemini}
Gemini, Rohan Anil, Sebastian Borgeaud, Yonghui Wu, Jean-Baptiste Alayrac, Jiahui Yu, Radu Soricut, Johan Schalkwyk, Andrew~M Dai, Anja Hauth, et~al. 2023.
\newblock Gemini: a family of highly capable multimodal models.
\newblock \emph{arXiv preprint arXiv:2312.11805}.

\bibitem[{GPT-4~Achiam et~al.(2023)GPT-4~Achiam, Adler, Agarwal, Ahmad, Akkaya, Aleman, Almeida, Altenschmidt, Altman, Anadkat et~al.}]{achiam2023gpt}
Josh GPT-4~Achiam, Steven Adler, Sandhini Agarwal, Lama Ahmad, Ilge Akkaya, Florencia~Leoni Aleman, Diogo Almeida, Janko Altenschmidt, Sam Altman, Shyamal Anadkat, et~al. 2023.
\newblock Gpt-4 technical report.
\newblock \emph{arXiv preprint arXiv:2303.08774}.

\bibitem[{Guo et~al.(2025)Guo, Yang, Zhang, Song, Zhang, Xu, Zhu, Ma, Wang, Bi et~al.}]{guo2025deepseek}
Daya Guo, Dejian Yang, Haowei Zhang, Junxiao Song, Ruoyu Zhang, Runxin Xu, Qihao Zhu, Shirong Ma, Peiyi Wang, Xiao Bi, et~al. 2025.
\newblock Deepseek-r1: Incentivizing reasoning capability in llms via reinforcement learning.
\newblock \emph{arXiv preprint arXiv:2501.12948}.

\bibitem[{Guo et~al.(2024)Guo, Zhu, Yang, Xie, Dong, Zhang, Chen, Bi, Wu, Li, Luo, Xiong, and Liang}]{deepseek-coder}
Daya Guo, Qihao Zhu, Dejian Yang, Zhenda Xie, Kai Dong, Wentao Zhang, Guanting Chen, Xiao Bi, Y.~Wu, Y.~K. Li, Fuli Luo, Yingfei Xiong, and Wenfeng Liang. 2024.
\newblock \href {https://arxiv.org/abs/2401.14196} {[link]}.

\bibitem[{Guo et~al.(2015)Guo, Zhang, Zhang, and Liu}]{guo2015efficient}
Qiang Guo, Caiming Zhang, Yunfeng Zhang, and Hui Liu. 2015.
\newblock An efficient svd-based method for image denoising.
\newblock \emph{IEEE transactions on Circuits and Systems for Video Technology}, 26(5):868--880.

\bibitem[{Huang and Wolkowicz(2018)}]{huang2018low}
Shimeng Huang and Henry Wolkowicz. 2018.
\newblock Low-rank matrix completion using nuclear norm minimization and facial reduction.
\newblock \emph{Journal of Global Optimization}, 72:5--26.

\bibitem[{Jiang et~al.(2023)Jiang, Sablayrolles, Mensch, Bamford, Chaplot, Casas, Bressand, Lengyel, Lample, Saulnier et~al.}]{jiang2023mistral}
Albert~Q Jiang, Alexandre Sablayrolles, Arthur Mensch, Chris Bamford, Devendra~Singh Chaplot, Diego de~las Casas, Florian Bressand, Gianna Lengyel, Guillaume Lample, Lucile Saulnier, et~al. 2023.
\newblock Mistral 7b.
\newblock \emph{arXiv preprint arXiv:2310.06825}.

\bibitem[{Kaplan et~al.(2020)Kaplan, McCandlish, Henighan, Brown, Chess, Child, Gray, Radford, Wu, and Amodei}]{kaplan2020scaling}
Jared Kaplan, Sam McCandlish, Tom Henighan, Tom~B Brown, Benjamin Chess, Rewon Child, Scott Gray, Alec Radford, Jeffrey Wu, and Dario Amodei. 2020.
\newblock Scaling laws for neural language models.
\newblock \emph{arXiv preprint arXiv:2001.08361}.

\bibitem[{Kung et~al.(1983)Kung, Arun, and Rao}]{kung1983state}
Sun-Yuan Kung, K~Si Arun, and DV~Bhaskar Rao. 1983.
\newblock State-space and singular-value decomposition-based approximation methods for the harmonic retrieval problem.
\newblock \emph{JOSA}, 73(12):1799--1811.

\bibitem[{Lian et~al.(2023{\natexlab{a}})Lian, Li, Yala, and Darrell}]{lian2023llm}
Long Lian, Boyi Li, Adam Yala, and Trevor Darrell. 2023{\natexlab{a}}.
\newblock Llm-grounded diffusion: Enhancing prompt understanding of text-to-image diffusion models with large language models.
\newblock \emph{arXiv preprint arXiv:2305.13655}.

\bibitem[{Lian et~al.(2023{\natexlab{b}})Lian, Goodson, Pentland et~al.}]{lian2023openorca}
W~Lian, B~Goodson, E~Pentland, et~al. 2023{\natexlab{b}}.
\newblock Openorca: An open dataset of gpt augmented flan reasoning traces.

\bibitem[{Lin(2004)}]{lin2004rouge}
Chin-Yew Lin. 2004.
\newblock Rouge: A package for automatic evaluation of summaries.
\newblock In \emph{Text summarization branches out}, pages 74--81.

\bibitem[{Liu et~al.(2023)Liu, Liu, Zhou, Ye, Chong, Zhou, Xie, Cao, Wang, You et~al.}]{liu2023llmrec}
Junling Liu, Chao Liu, Peilin Zhou, Qichen Ye, Dading Chong, Kang Zhou, Yueqi Xie, Yuwei Cao, Shoujin Wang, Chenyu You, et~al. 2023.
\newblock Llmrec: Benchmarking large language models on recommendation task.
\newblock \emph{arXiv preprint arXiv:2308.12241}.

\bibitem[{Mihaylov et~al.(2018)Mihaylov, Clark, Khot, and Sabharwal}]{mihaylov2018can}
Todor Mihaylov, Peter Clark, Tushar Khot, and Ashish Sabharwal. 2018.
\newblock Can a suit of armor conduct electricity? a new dataset for open book question answering.
\newblock \emph{arXiv preprint arXiv:1809.02789}.

\bibitem[{Neubig(2017)}]{neubig2017neural}
Graham Neubig. 2017.
\newblock Neural machine translation and sequence-to-sequence models: A tutorial.
\newblock \emph{arXiv preprint arXiv:1703.01619}.

\bibitem[{Papineni et~al.(2002)Papineni, Roukos, Ward, and Zhu}]{papineni2002bleu}
Kishore Papineni, Salim Roukos, Todd Ward, and Wei-Jing Zhu. 2002.
\newblock Bleu: a method for automatic evaluation of machine translation.
\newblock In \emph{Proceedings of the 40th annual meeting of the Association for Computational Linguistics}, pages 311--318.

\bibitem[{Ruan et~al.(2024)Ruan, Maddison, and Hashimoto}]{ruan2024observational}
Yangjun Ruan, Chris~J Maddison, and Tatsunori Hashimoto. 2024.
\newblock Observational scaling laws and the predictability of language model performance.
\newblock \emph{arXiv preprint arXiv:2405.10938}.

\bibitem[{Sakaguchi et~al.(2021)Sakaguchi, Bras, Bhagavatula, and Choi}]{sakaguchi2021winogrande}
Keisuke Sakaguchi, Ronan~Le Bras, Chandra Bhagavatula, and Yejin Choi. 2021.
\newblock Winogrande: An adversarial winograd schema challenge at scale.
\newblock \emph{Communications of the ACM}, 64(9):99--106.

\bibitem[{Sasaki(2007)}]{2007The}
Yutaka Sasaki. 2007.
\newblock The truth of the f-measure.
\newblock \emph{Teach tutor mater}.

\bibitem[{Saul et~al.(2005)Saul, Weiss, and Bottou}]{saul2005advances}
Lawrence~K Saul, Yair Weiss, and L{\'e}on Bottou. 2005.
\newblock \emph{Advances in neural information processing systems 17: proceedings of the 2004 conference}, volume~17.
\newblock MIT Press.

\bibitem[{Sawada et~al.(2023)Sawada, Paleka, Havrilla, Tadepalli, Vidas, Kranias, Nay, Gupta, and Komatsuzaki}]{sawada2023arb}
Tomohiro Sawada, Daniel Paleka, Alexander Havrilla, Pranav Tadepalli, Paula Vidas, Alexander Kranias, John~J Nay, Kshitij Gupta, and Aran Komatsuzaki. 2023.
\newblock Arb: Advanced reasoning benchmark for large language models.
\newblock \emph{arXiv preprint arXiv:2307.13692}.

\bibitem[{Shannon(1948)}]{shannon1948mathematical}
Claude~Elwood Shannon. 1948.
\newblock A mathematical theory of communication.
\newblock \emph{The Bell system technical journal}, 27(3):379--423.

\bibitem[{{Skylion007}(2019)}]{openwebtextcorpus}
{Skylion007}. 2019.
\newblock {OpenWebText Corpus}.
\newblock {\url{http://Skylion007.github.io/OpenWebTextCorpus}}.
\newblock [Online; accessed 2024-09-27].

\bibitem[{Team et~al.(2024)Team, Mesnard, Hardin, Dadashi, Bhupatiraju, Pathak, Sifre, Rivi{\`e}re, Kale, Love et~al.}]{team2024gemma}
Gemma Team, Thomas Mesnard, Cassidy Hardin, Robert Dadashi, Surya Bhupatiraju, Shreya Pathak, Laurent Sifre, Morgane Rivi{\`e}re, Mihir~Sanjay Kale, Juliette Love, et~al. 2024.
\newblock Gemma: Open models based on gemini research and technology.
\newblock \emph{arXiv preprint arXiv:2403.08295}.

\bibitem[{Wang et~al.(2024)Wang, Chen, Chen, Wu, Zhu, Zeng, Luo, Lu, Zhou, Qiao et~al.}]{wang2024visionllm}
Wenhai Wang, Zhe Chen, Xiaokang Chen, Jiannan Wu, Xizhou Zhu, Gang Zeng, Ping Luo, Tong Lu, Jie Zhou, Yu~Qiao, et~al. 2024.
\newblock Visionllm: Large language model is also an open-ended decoder for vision-centric tasks.
\newblock \emph{Advances in Neural Information Processing Systems}, 36.

\bibitem[{Wei et~al.(2024)Wei, Tan, Li, Wang, and Huang}]{wei2024large}
Lai Wei, Zhiquan Tan, Chenghai Li, Jindong Wang, and Weiran Huang. 2024.
\newblock Large language model evaluation via matrix entropy.
\newblock \emph{arXiv preprint arXiv:2401.17139}.

\bibitem[{Yang et~al.(2024)Yang, Yang, Hui, Zheng, Yu, Zhou, Li, Li, Liu, Huang et~al.}]{yang2024qwen2}
An~Yang, Baosong Yang, Binyuan Hui, Bo~Zheng, Bowen Yu, Chang Zhou, Chengpeng Li, Chengyuan Li, Dayiheng Liu, Fei Huang, et~al. 2024.
\newblock Qwen2 technical report.
\newblock \emph{arXiv preprint arXiv:2407.10671}.

\bibitem[{Zhang et~al.(2024)Zhang, Ladhak, Durmus, Liang, McKeown, and Hashimoto}]{zhang2024benchmarking}
Tianyi Zhang, Faisal Ladhak, Esin Durmus, Percy Liang, Kathleen McKeown, and Tatsunori~B Hashimoto. 2024.
\newblock Benchmarking large language models for news summarization.
\newblock \emph{Transactions of the Association for Computational Linguistics}, 12:39--57.

\bibitem[{Zhang(2015)}]{zhang2015singular}
Zhihua Zhang. 2015.
\newblock The singular value decomposition, applications and beyond.
\newblock \emph{arXiv preprint arXiv:1510.08532}.

\bibitem[{Zhao et~al.(2023)Zhao, Zhou, Li, Tang, Wang, Hou, Min, Zhang, Zhang, Dong et~al.}]{zhao2023survey}
Wayne~Xin Zhao, Kun Zhou, Junyi Li, Tianyi Tang, Xiaolei Wang, Yupeng Hou, Yingqian Min, Beichen Zhang, Junjie Zhang, Zican Dong, et~al. 2023.
\newblock A survey of large language models.
\newblock \emph{arXiv preprint arXiv:2303.18223}.

\bibitem[{Zheng et~al.(2023)Zheng, Chiang, Sheng, Zhuang, Wu, Zhuang, Lin, Li, Li, Xing et~al.}]{zheng2023judging}
Lianmin Zheng, Wei-Lin Chiang, Ying Sheng, Siyuan Zhuang, Zhanghao Wu, Yonghao Zhuang, Zi~Lin, Zhuohan Li, Dacheng Li, Eric Xing, et~al. 2023.
\newblock Judging llm-as-a-judge with mt-bench and chatbot arena.
\newblock \emph{Advances in Neural Information Processing Systems}, 36:46595--46623.

\end{thebibliography}

\newpage
\appendix
\section{Appendix}
\subsection{Ablation Study}
\label{sec:ablation-study}
To thoroughly validate the rationale behind our metric design, experimental framework, and the efficacy of Matrix Nuclear-Norm, we conducted a series of ablation studies.

\subsubsection{Different Model Family}
\label{sec:model-dataset}
\begin{figure}[ht]
    \centering
    \begin{subfigure}[b]{0.45\textwidth}
        \includegraphics[width=\textwidth]{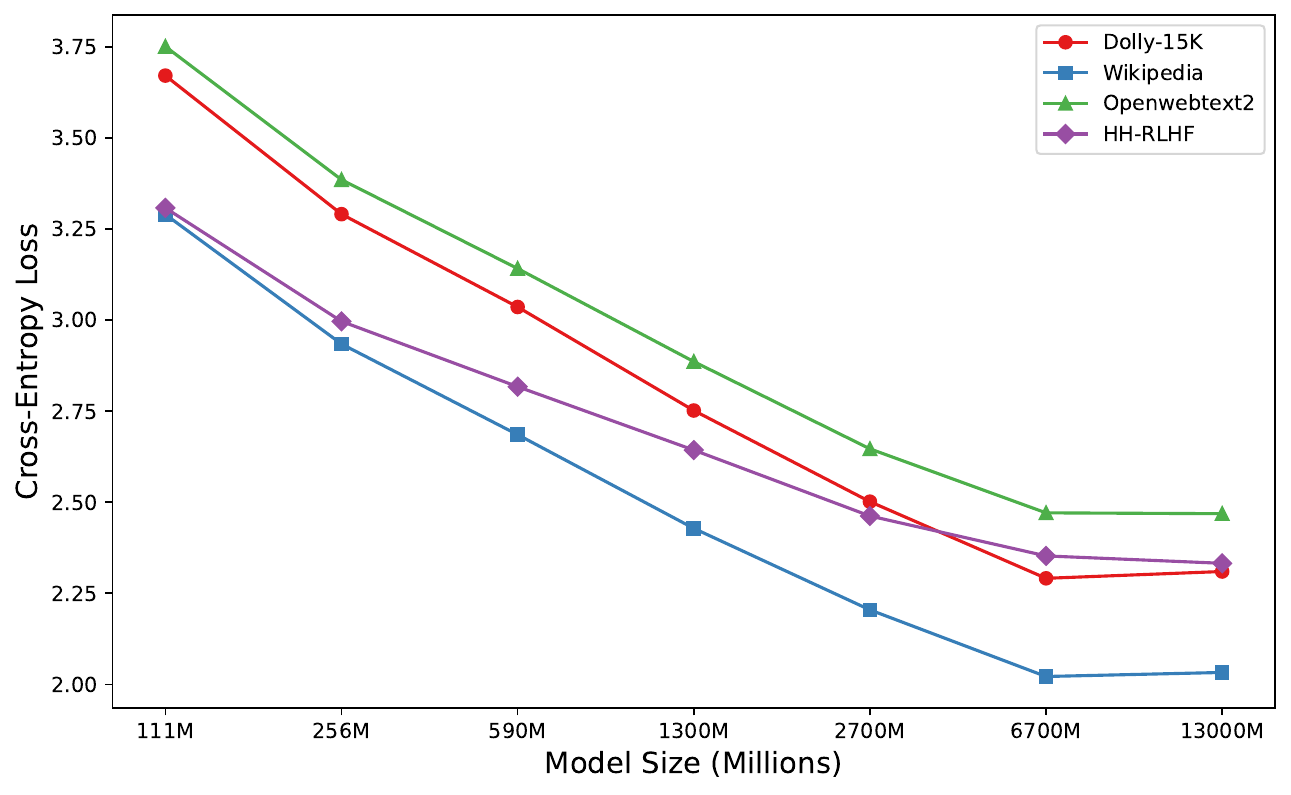} 
        \caption{Cross-Entropy Loss}
        \label{fig:loss-gpt-fig}
    \end{subfigure}
    \begin{subfigure}[b]{0.45\textwidth}
        \includegraphics[width=\textwidth]{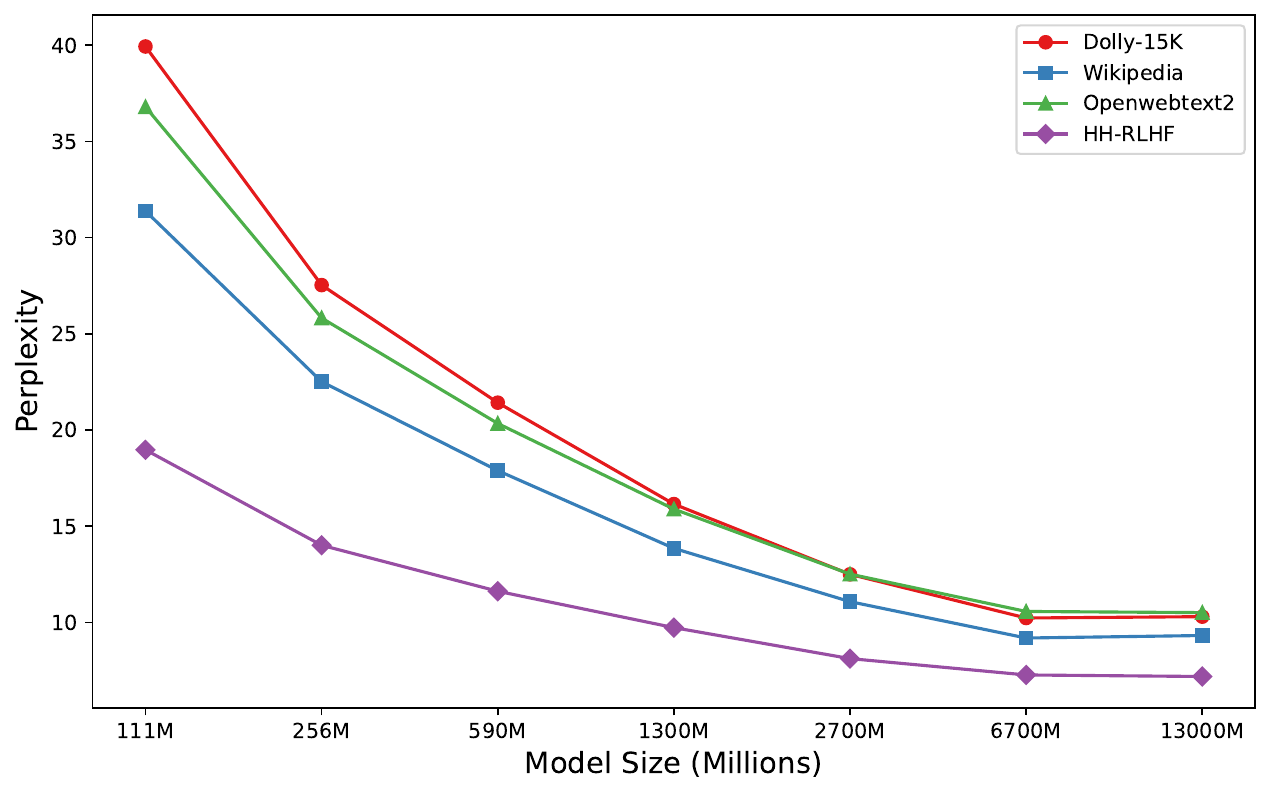} 
        \caption{Perplexity}
        \label{fig:perplexity-gpt-fig}
    \end{subfigure}
    \vspace{1em} 
    \caption{Comparison of loss, and perplexity when model scales up.} 
\end{figure}

\begin{figure}[ht]
    \centering
    \begin{subfigure}[b]{0.4\textwidth}
        \includegraphics[width=\textwidth]{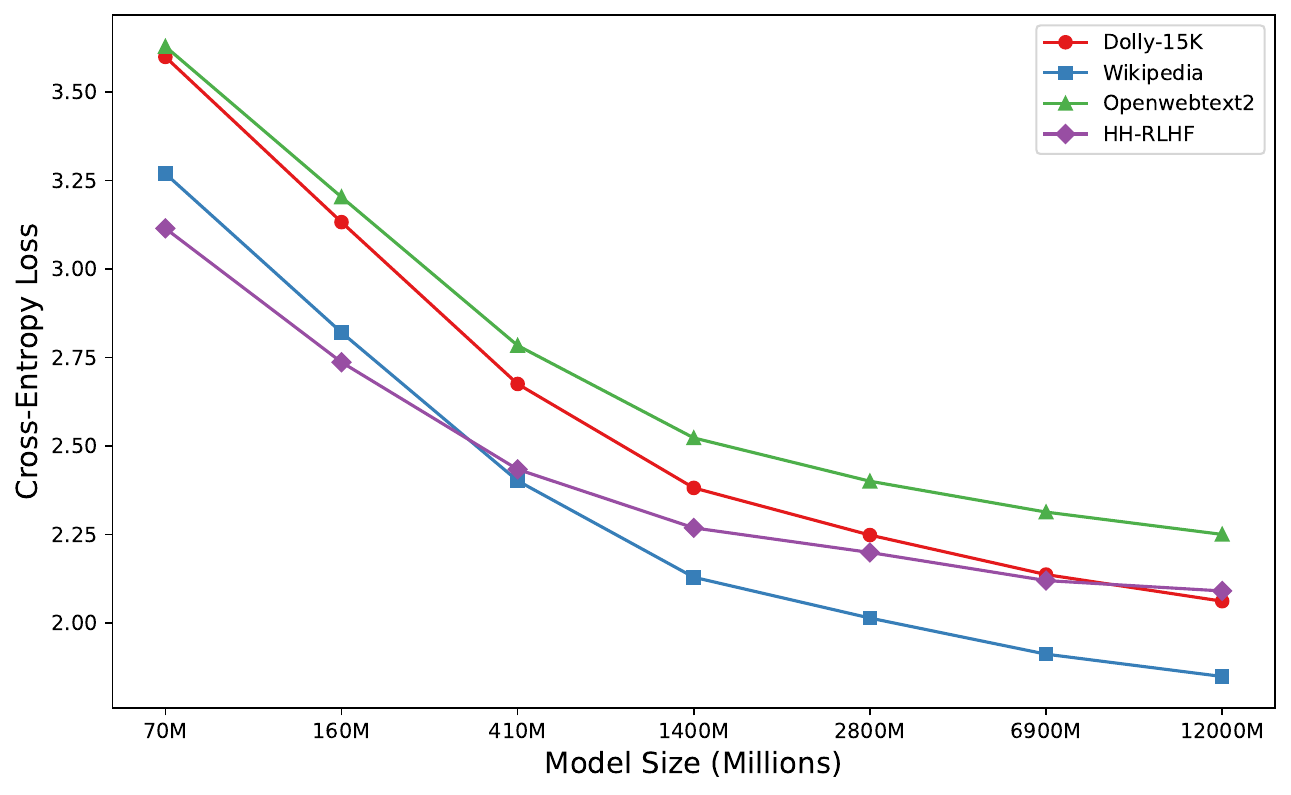} 
        \caption{Cross-Entropy Loss}
        \label{fig:loss-pythia-fig}
    \end{subfigure}
    \begin{subfigure}[b]{0.4\textwidth}
        \includegraphics[width=\textwidth]{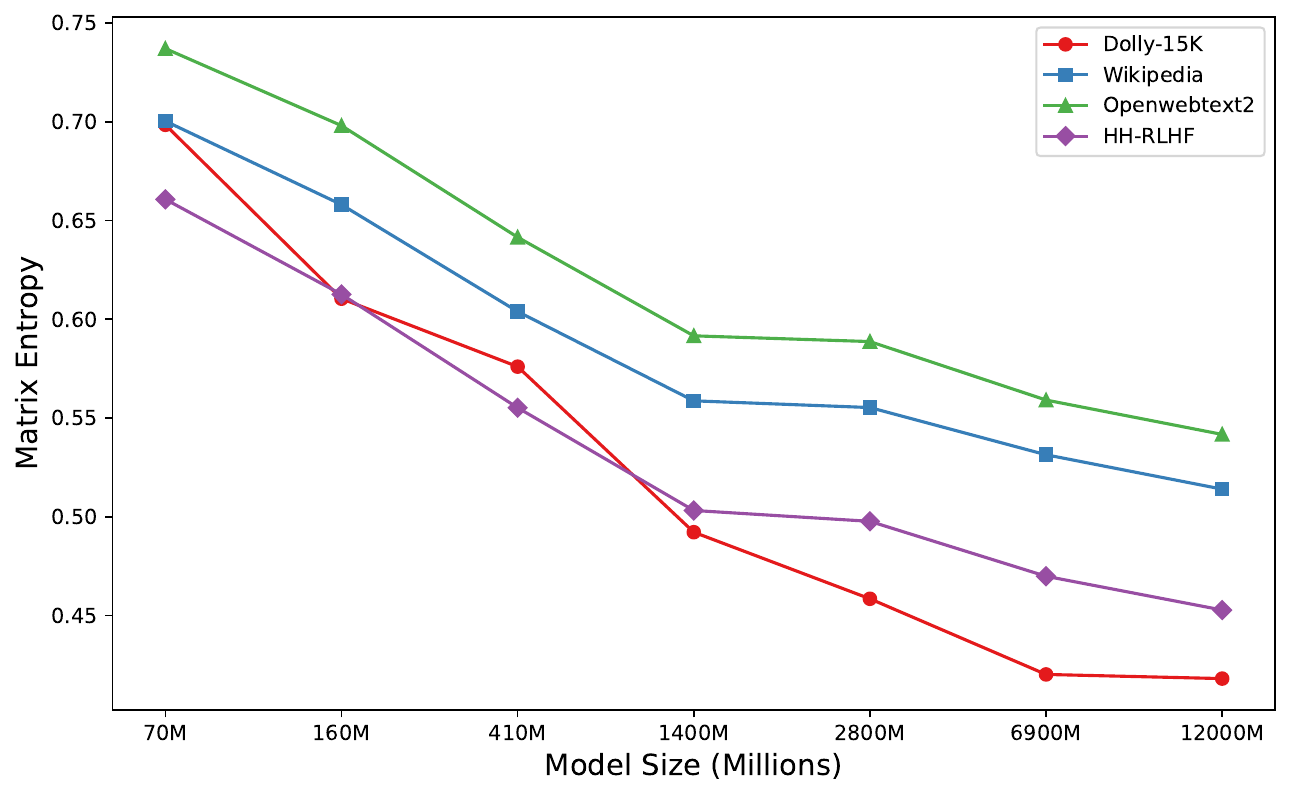} 
        \caption{Matrix Entropy}
        \label{fig:matrix-entropy-pythia-fig}
    \end{subfigure}
    \begin{subfigure}[b]{0.4\textwidth}
        \includegraphics[width=\textwidth]{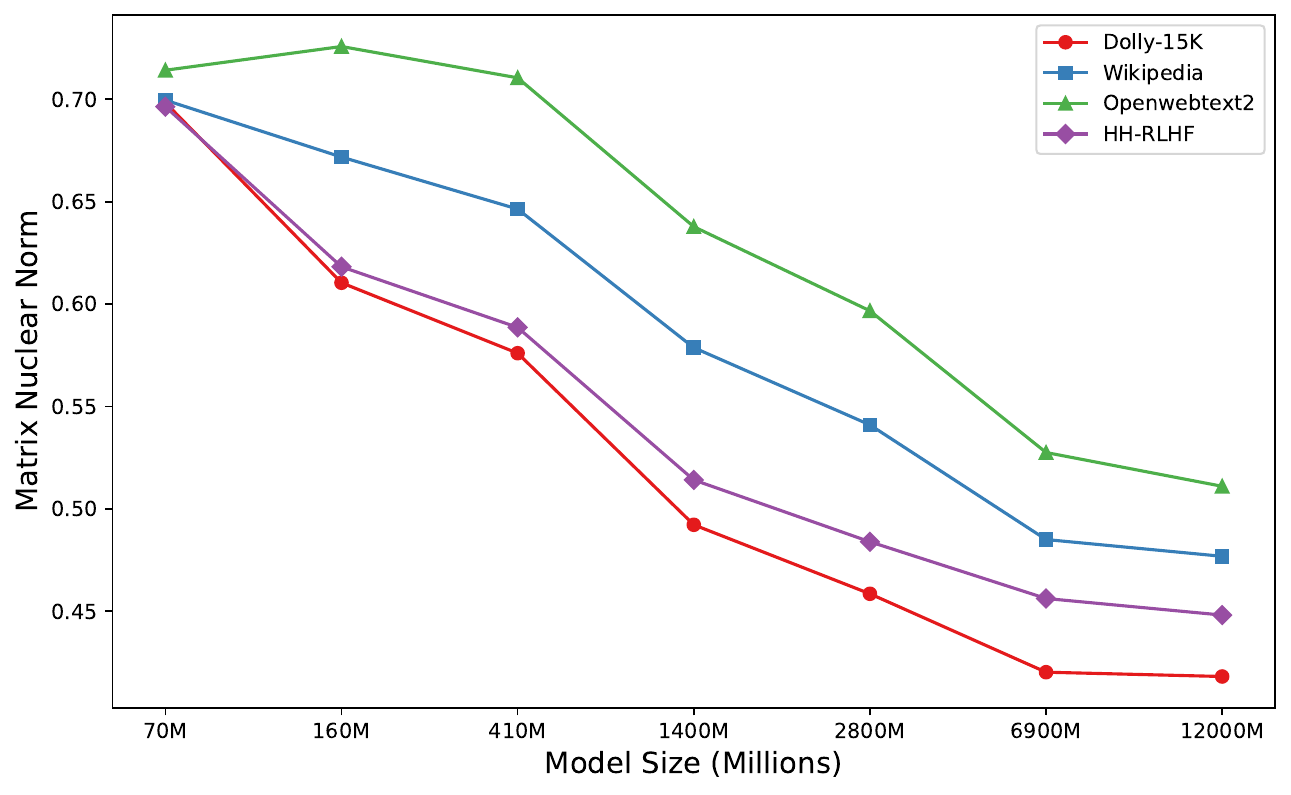} 
        \caption{Matrix Nuclear-Norm}
        \label{fig:Matrix Nuclear-Norm-pythia-fig}
    \end{subfigure}
    \hfill 
    \caption{Pythia Model Metrics: Matrix Nuclear-Norm, Matrix Entropy, and Loss} 
    \label{fig:Pythia Model Metrics-pythia-fig}
\end{figure}

In addition to evaluating Matrix Nuclear-Norm within the Cerebras-GPT model series, we extended our experiments to the Pythia model family, which spans from 14M to 12B parameters and is trained on consistent public datasets. Utilizing the same datasets as described in Section \ref{sec:Scaling Law of Matrix Nuclear-Norm}, we computed matrix entropy, loss values, and Matrix Nuclear-Norm for these models. The empirical results (see Figure \ref{fig:Matrix Nuclear-Norm-pythia-fig}) demonstrate that the Matrix Nuclear-Norm values for the Pythia models adhere to established scaling laws. However, we excluded metrics for the 14M, 31M, and 1B models due to notable deviations from the expected range, likely stemming from the inherent instability associated with smaller parameter sizes when tackling complex tasks. This further reinforces Matrix Nuclear-Norm as a robust metric for assessing model performance, underscoring its utility in the comparative analysis of LLMs.

Moreover, we compared the computation times for Matrix Entropy and Matrix Nuclear-Norm across the Pythia models (can see in Figure \ref{tab:time-pythia}). The results unequivocally indicate that Matrix Nuclear-Norm necessitates considerably less computation time than Matrix Entropy, underscoring its efficiency. Detailed results are summarized in Table \ref{tab:four-dataset-pythia}.

\subsubsection{Sampling Strategy}

In the ablation experiments, we extracted a baseline subset of 10,000 entries from the extensive Wikipedia dataset using three random seeds to evaluate the robustness of the Matrix Nuclear-Norm metric. We also tested additional subsets of 15,000 and 20,000 entries due to potential entry count issues. Given the large scale of the datasets, comprehensive calculations were impractical, so we employed random sampling.

The results showed that variations in random seeds and sample sizes had minimal impact on Matrix Nuclear-Norm values, with a standard deviation of only 0.0004975 (see Table \ref{tab:sampling-strategy}), indicating high consistency across trials. These findings confirm the Matrix Nuclear-Norm as a reliable metric for large-scale datasets, effectively evaluating information compression and redundancy elimination in LLMs.


\begin{table*}[ht]
\caption{Ablation study of different sampling strategies on the Wikimedia\citep{wikimedia_downloads} dataset.}
\label{tab:sampling-strategy}
\resizebox{\linewidth}{!}{ 
\begin{tabular}{c|ccccc|c}
\toprule
\multirow{2}{*}{\textbf{MODEL}} & \multicolumn{5}{c|}{\textbf{SAMPLING STRATEGY}}                                                               & \multirow{2}{*}{\textbf{STANDARD DEVIATION}} \\
                       & \textbf{10000 (SEED 1)} & \textbf{10000 (SEED 2)} & \textbf{10000 (SEED 3)} & \textbf{15000} & \textbf{20000} &                                     \\ \midrule
CEREBRAS-GPT-1.3B      & 0.5684         & 0.5670                  & 0.5676                  & 0.5699         & 0.5693         & 0.0004975                           \\ \bottomrule
\end{tabular}
}
\end{table*}

\subsection{Model Selection and Datasets for Analysis}
\label{sec:model and dataset}
\textbf{{Model Selection.}}
To investigate language model scaling, we employed a diverse set of transformer-based large language models (LLMs) across varying parameter sizes. A key focus of our analysis was the Cerebras-GPT model \citep{gao2020pile}, which ranges from 111 million to 13 billion parameters, providing a comprehensive look at scaling effects in pre-trained models. Additionally, we included scaled versions of the Pythia model \citep{biderman2023pythia}, with parameter counts ranging from 14 million to 12 billion, enabling a broader analysis of model performance across different scales.

To ensure a well-rounded evaluation, we also tested a variety of models, including the DeepSeek series (1.3B, 6.7B, 7B) \citep{deepseek-coder}, Llama3 series (8B, 70B) \citep{dubey2024llama}, QWEN 2 series (0.5B, 1.5B, 7B, 72B) \citep{yang2024qwen2}, and Vicuna models (7B, 13B, 33B) \citep{chiang2023vicuna}. For additional comparative insights, we included models of similar scale, such as Gemma-7B \citep{team2024gemma} and Mistral-7B \citep{jiang2023mistral}.

\textbf{Datasets for Analysis.}
Our experiments were conducted using several key benchmark datasets. We selected AlpacaEval\citep{dubois2024length} and ChatBot Arena \citep{zheng2023judging} as the primary datasets for model evaluation. Additionally, subsets from Wikipedia \citep{wikimedia_downloads} and OpenWebText2 \citep{openwebtextcorpus} were utilized to track variations in Matrix Nuclear-Norm values, especially with the Cerebras-GPT models.

To validate the Matrix Nuclear-Norm metric, we employed the dolly-15k dataset \citep{conover2023free} for instruction tuning and the hh-rlhf dataset \citep{bai2022training} for reinforcement learning with human feedback (RLHF). Further evaluations were performed on benchmark datasets such as OpenBookQA \citep{mihaylov2018can}, Winogrande \citep{sakaguchi2021winogrande}, and PIQA \citep{bisk2020piqa}. Lastly, prompt learning experiments with the OpenOrca dataset \citep{lian2023openorca} provided a comprehensive framework for assessing the Matrix Nuclear-Norm's effectiveness across a variety of inference tasks.

\subsection{Supplementary Experiment Results}
\begin{figure}[ht]
    \centering
    \includegraphics[width=0.5\textwidth]{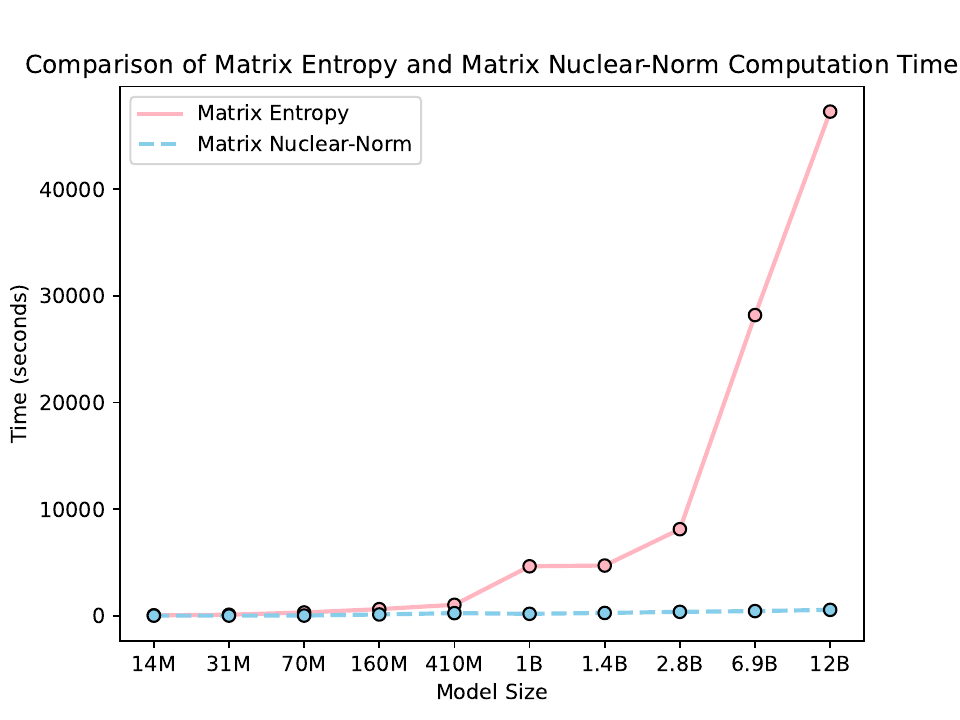} 
    \caption{Pythia: Time Comparison of Matrix Entropy and Nuclear-Norm}
    \label{fig:time-pythia-fig}
\end{figure}

The following results provide additional insights into the Matrix Nuclear-Norm evaluations and comparisons across various language models:
\begin{enumerate}
    \item Tables \ref{tab:Matrix Nuclear-Norm of Llama-3} and \ref{tab:Matrix Nuclear-Norm of QWEN 2} present the Matrix Nuclear-Norm evaluation results during the inference process for Llama-3 and QWEN-2.
    \item Figure \ref{fig:time-pythia-fig} illustrates that as model size increases, the computation time for Matrix Entropy grows exponentially, while Matrix Nuclear-Norm demonstrates a significant time advantage. This further emphasizes Matrix Nuclear-Norm's efficiency in assessing model performance.The complete results are presented in Table \ref{tab:time-pythia}, which includes all relevant time data for the Pythia model family.
    \item Table \ref{tab:four-dataset-gpt} contains the complete results for the comparison of Matrix Nuclear-Norm and other metrics based on Cerebras-GPT  family considered in Figure \ref{fig:Matrix Nuclear-Norm-gpt-fig}.
    \item Table \ref{tab:Three-dataset-gpt} demonstrates the correlation between Matrix Nuclear-Norm and other benchmark indicators, showing a consistent trend where values decrease as model size increases. This analysis examines the performance of language modeling indicators across OpenBookQA, Winogrande, and PIQA datasets.

    \item Table \ref{tab:four-dataset-pythia} illustrates the numerical results of Figure \ref{fig:Matrix Nuclear-Norm-pythia-fig} in the ablation study of Pythia family.
    \item Table \ref{tab:three-prompt} shows the prompts used for the investigation of prompt learning.
\end{enumerate}

\begin{table}
    \resizebox{\linewidth}{!}{ 
    \begin{tabular}{@{}
    >{\columncolor[HTML]{FFFFFF}}c 
    >{\columncolor[HTML]{FFFFFF}}c 
    >{\columncolor[HTML]{FFFFFF}}c 
    >{\columncolor[HTML]{FFFFFF}}c 
    >{\columncolor[HTML]{FFFFFF}}c 
    >{\columncolor[HTML]{FFFFFF}}c @{}}
    \toprule
    {\textbf{Model}}                                   & {\textbf{DataSet}} & {\textbf{0.5B}} & {\textbf{1.5B}} & {\textbf{7B}} & {\textbf{72B}} \\ \midrule
    \cellcolor[HTML]{FFFFFF}{}                         & {Alpaca}           & {0.6551}        & {0.6176}        & {0.5989}      & {0.5261}       \\
    \multirow{-2}{*}{\cellcolor[HTML]{FFFFFF}{QWEN 2}} & {Arena}            & {0.6872}        & {0.6374}        & {0.5751}      & {0.4689}       \\ \bottomrule
    \end{tabular}
    }
    \caption{Matrix Nuclear-Norm in QWEN 2 Responses}
    \label{tab:Matrix Nuclear-Norm of QWEN 2}
\end{table}

\begin{table}
\centering
\resizebox{0.68\linewidth}{!}{ 
\begin{tabular}{@{}ccccc@{}}
\toprule
\textbf{Model} &{\textbf{DataSet}} & \textbf{8B} & \textbf{70B} \\ \midrule
\multirow{2}{*}{Llama-3}  & {Alpaca} & 0.5782      & 0.4935       \\
& {Arena} & 0.5817      & 0.4967       \\ \bottomrule
\end{tabular}
}
\caption{Matrix Nuclear-Norm in Llama 3 Responses}
\label{tab:Matrix Nuclear-Norm of Llama-3}
\end{table}

\begin{table}[ht]
\centering
\begin{tabular}{@{}cccc@{}}
\toprule
\textbf{Model Size} & \textbf{ME (s)} & \textbf{MNN (s)} & \textbf{Ratio } \\ \midrule
14M                 & 52.9                             & 22.3                    & 2.4                     \\
31M                 & 114.1                            & 28.2                    & 4.0                     \\
70M                 & 320.7                            & 24.3                    & 13.2                    \\
160M                & 632.0                            & 41.6                    & 15.2                    \\
410M                & 1040.9                           & 81.0                    & 12.8                    \\
1B                  & 4650.1                           & 114.1                   & 40.8                    \\
1.4B                & 6387.0                           & 347.9                   & 18.4                    \\
2.8B                & 8127.1                           & 343.4                   & 23.7                    \\
6.9B                & 28197.8                          & 816.6                   & 34.5                    \\
12B                 & 47273.5                          & 1276.1                  & \textbf{37.0}                    \\ \bottomrule
\end{tabular}
\caption{Pythia Model: Matrix Entropy(ME) vs. Matrix Nuclear-Norm(MNN) Time Comparison}
\label{tab:time-pythia}
\end{table}

\begin{table*}[ht]
\resizebox{\linewidth}{!}{
\begin{tabular}{@{}
>{\columncolor[HTML]{FFFFFF}}c 
>{\columncolor[HTML]{FFFFFF}}c 
>{\columncolor[HTML]{FFFFFF}}c 
>{\columncolor[HTML]{FFFFFF}}c 
>{\columncolor[HTML]{FFFFFF}}c 
>{\columncolor[HTML]{FFFFFF}}c 
>{\columncolor[HTML]{FFFFFF}}c 
>{\columncolor[HTML]{FFFFFF}}c 
>{\columncolor[HTML]{FFFFFF}}c @{}}
\toprule
\cellcolor[HTML]{FFFFFF}                                                    & \cellcolor[HTML]{FFFFFF}                                      & \multicolumn{7}{c}{\cellcolor[HTML]{FFFFFF}GPT MODEL SIZE}                                                                                                                                                                                                                         \\
\multirow{-2}{*}{\cellcolor[HTML]{FFFFFF}BENCHMARKS}                        & \multirow{-2}{*}{\cellcolor[HTML]{FFFFFF}INDICATORS}          & {\color[HTML]{333333} \textbf{111M}} & {\color[HTML]{333333} \textbf{256M}} & {\color[HTML]{333333} \textbf{590M}} & {\color[HTML]{333333} \textbf{1.3B}} & {\color[HTML]{333333} \textbf{2.7B}}   & {\color[HTML]{333333} \textbf{6.7B}} & {\color[HTML]{333333} \textbf{13B}}    \\ \midrule
\cellcolor[HTML]{FFFFFF}{\color[HTML]{333333} }                             & {\color[HTML]{333333} ACCURACY}                               & {\color[HTML]{333333} 0.118}         & {\color[HTML]{333333} 0.158}         & {\color[HTML]{333333} 0.158}         & {\color[HTML]{333333} 0.166}         & {\color[HTML]{333333} 0.206}           & {\color[HTML]{333333} 0.238}         & {\color[HTML]{333333} 0.286}           \\
\cellcolor[HTML]{FFFFFF}{\color[HTML]{333333} }                             & {\color[HTML]{333333} MATRIX ENTROPY}                         & {\color[HTML]{333333} 0.3575}        & {\color[HTML]{333333} 0.3416}        & {\color[HTML]{333333} 0.3237}        & {\color[HTML]{333333} 0.3140}        & {\color[HTML]{333333} 0.2991}          & {\color[HTML]{333333} 0.2848}        & {\color[HTML]{333333} 0.2767}          \\
\cellcolor[HTML]{FFFFFF}{\color[HTML]{333333} }                             & {\color[HTML]{333333} LOSS}                                   & {\color[HTML]{333333} 5.6196}        & {\color[HTML]{333333} 5.3536}        & {\color[HTML]{333333} 5.1881}        & {\color[HTML]{333333} 4.9690}        & {\color[HTML]{333333} 4.8723}          & {\color[HTML]{333333} 4.7195}        & {\color[HTML]{333333} 4.7050}          \\
\cellcolor[HTML]{FFFFFF}{\color[HTML]{333333} }                             & {\color[HTML]{333333} PPL}                                    & {\color[HTML]{333333} 148.38}        & {\color[HTML]{333333} 108.10}        & {\color[HTML]{333333} 83.45}         & {\color[HTML]{333333} 65.10}         & {\color[HTML]{333333} 50.93}           & {\color[HTML]{333333} 41.80}         & {\color[HTML]{333333} 40.89}           \\
\multirow{-5}{*}{\cellcolor[HTML]{FFFFFF}{\color[HTML]{333333} OPENBOOKQA}} & {\color[HTML]{333333} MATRIX NUCLEAR-NORM}                         & {\color[HTML]{333333} 0.4447}        & {\color[HTML]{333333} 0.4057}        & {\color[HTML]{333333} 0.3941}        & {\color[HTML]{333333} 0.3644}        & {\color[HTML]{FF0000} \textbf{0.4606}} & {\color[HTML]{333333} 0.3672}        & {\color[HTML]{FF0000} \textbf{0.4423}} \\ \midrule
\cellcolor[HTML]{FFFFFF}{\color[HTML]{333333} }                             & \cellcolor[HTML]{FFFFFF}{\color[HTML]{333333} ACCURACY}       & {\color[HTML]{333333} 0.488}         & {\color[HTML]{333333} 0.511}         & {\color[HTML]{333333} 0.498}         & {\color[HTML]{333333} 0.521}         & {\color[HTML]{333333} 0.559}           & {\color[HTML]{333333} 0.602}         & {\color[HTML]{333333} 0.646}           \\
\cellcolor[HTML]{FFFFFF}{\color[HTML]{333333} }                             & \cellcolor[HTML]{FFFFFF}{\color[HTML]{333333} MATRIX ENTROPY} & {\color[HTML]{333333} 0.4073}        & {\color[HTML]{333333} 0.3915}        & {\color[HTML]{333333} 0.3706}        & {\color[HTML]{333333} 0.3605}        & {\color[HTML]{333333} 0.3419}          & {\color[HTML]{333333} 0.3272}        & {\color[HTML]{333333} 0.3149}          \\
\cellcolor[HTML]{FFFFFF}{\color[HTML]{333333} }                             & \cellcolor[HTML]{FFFFFF}{\color[HTML]{333333} LOSS}           & {\color[HTML]{333333} 4.7869}        & {\color[HTML]{333333} 4.5854}        & {\color[HTML]{333333} 4.4141}        & {\color[HTML]{333333} 4.2513}        & {\color[HTML]{333333} 4.1107}          & {\color[HTML]{333333} 4.0109}        & {\color[HTML]{333333} 4.0266}          \\
\cellcolor[HTML]{FFFFFF}{\color[HTML]{333333} }                             & {\color[HTML]{333333} PPL}                                    & {\color[HTML]{333333} 39.81}         & {\color[HTML]{333333} 30.25}         & {\color[HTML]{333333} 26.57}         & {\color[HTML]{333333} 21.87}         & {\color[HTML]{333333} 18.55}           & {\color[HTML]{333333} 16.53}         & {\color[HTML]{333333} 16.94}           \\
\multirow{-5}{*}{\cellcolor[HTML]{FFFFFF}{\color[HTML]{333333} WINOGRANDE}} & {\color[HTML]{333333} MATRIX NUCLEAR-NORM}                         & {\color[HTML]{333333} 0.4802}        & {\color[HTML]{333333} 0.4479}        & {\color[HTML]{333333} 0.4440}        & {\color[HTML]{333333} 0.4133}        & {\color[HTML]{FF0000} \textbf{0.5232}} & {\color[HTML]{333333} 0.4220}        & {\color[HTML]{FF0000} \textbf{0.4964}} \\ \midrule
\cellcolor[HTML]{FFFFFF}{\color[HTML]{333333} }                             & \cellcolor[HTML]{FFFFFF}{\color[HTML]{333333} ACCURACY}       & {\color[HTML]{333333} 0.594}         & {\color[HTML]{333333} 0.613}         & {\color[HTML]{333333} 0.627}         & {\color[HTML]{333333} 0.664}         & {\color[HTML]{333333} \textbf{0.701}}  & {\color[HTML]{333333} 0.739}         & {\color[HTML]{333333} \textbf{0.766}}  \\
\cellcolor[HTML]{FFFFFF}{\color[HTML]{333333} }                             & \cellcolor[HTML]{FFFFFF}{\color[HTML]{333333} MATRIX ENTROPY} & {\color[HTML]{333333} 0.4168}        & {\color[HTML]{333333} 0.3991}        & {\color[HTML]{333333} 0.3783}        & {\color[HTML]{333333} 0.3676}        & {\color[HTML]{333333} 0.3504}          & {\color[HTML]{333333} 0.3344}        & {\color[HTML]{333333} 0.3264}          \\
\cellcolor[HTML]{FFFFFF}{\color[HTML]{333333} }                             & \cellcolor[HTML]{FFFFFF}{\color[HTML]{333333} LOSS}           & {\color[HTML]{333333} 4.8425}        & {\color[HTML]{333333} 4.5470}        & {\color[HTML]{333333} 4.4029}        & {\color[HTML]{333333} 4.1613}        & {\color[HTML]{333333} 4.0075}          & {\color[HTML]{333333} 3.8545}        & {\color[HTML]{333333} 3.8826}          \\
\cellcolor[HTML]{FFFFFF}{\color[HTML]{333333} }                             & {\color[HTML]{333333} PPL}                                    & {\color[HTML]{333333} 69.80}         & {\color[HTML]{333333} 47.94}         & {\color[HTML]{333333} 37.88}         & {\color[HTML]{333333} 28.76}         & {\color[HTML]{333333} 23.15}           & {\color[HTML]{333333} 19.76}         & {\color[HTML]{333333} 19.72}           \\
\multirow{-5}{*}{\cellcolor[HTML]{FFFFFF}{\color[HTML]{333333} PIQA}}       & {\color[HTML]{333333} MATRIX NUCLEAR-NORM}                         & {\color[HTML]{333333} 0.4868}        & {\color[HTML]{333333} 0.4327}        & {\color[HTML]{333333} 0.4164}        & {\color[HTML]{333333} 0.3826}        & {\color[HTML]{FF0000} \textbf{0.4452}} & {\color[HTML]{333333} 0.3675}        & {\color[HTML]{FF0000} \textbf{0.4149}} \\ \bottomrule
\end{tabular}
}
\caption{Language modeling indicators on openbookqa, winogrande and piqa.Except for the matrix nuclear norm, the data is sourced from \cite{wei2024large}}
\label{tab:Three-dataset-gpt}
\end{table*}

\renewcommand{\arraystretch}{1} 
\begin{table*}[ht]
\caption{The table illustrates the performance metrics for a range of GPT models on the Dolly-15k, Wikipedia, OpenWebText2, and HH-RLHF datasets, encompassing matrix entropy, loss, and perplexity. Except for the matrix nuclear norm, the data is sourced from \cite{wei2024large}, underscoring the relationship between model scale and its performance.
}
\label{tab:four-dataset-gpt}
\resizebox{\linewidth}{!}{
\begin{tabular}{ccccccccc}
\midrule
\multirow{2}{*}{DATASET}     & \multirow{2}{*}{INDICATORS} & \multicolumn{7}{c}{GPT MODELS SIZE}                          \\
                              &                             & 111M   & 256M   & 590M   & 1.3B   & 2.7B   & 6.7B   & 13B    \\ \midrule
\multirow{4}{*}{DOLLY-15K}    & MATRIX ENTROPY              & 0.5976 & 0.5840 & 0.5582 & 0.5477 & 0.5240 & 0.5064 & 0.4859 \\
                              & LOSS                        & 3.6710 & 3.2907 & 3.0359 & 2.7517 & 2.5015 & 2.2911 & 2.3098 \\
                              & PPL                         & 39.93  & 27.53  & 21.42  & 16.15  & 12.50  & 10.23  & 10.30  \\
                              & MATRIX NUCLEAR-NORM                       & 0.6207 & 0.5565 & 0.5063 & 0.4553 & 0.4639 & 0.3904 & 0.4859 \\ \midrule
\multirow{4}{*}{WIKIPEDIA}    & MATRIX ENTROPY              & 0.6177 & 0.6077 & 0.5848 & 0.5786 & 0.5523 & 0.5368 & 0.5126 \\
                              & LOSS                        & 3.2900 & 2.9343 & 2.6854 & 2.4282 & 2.2045 & 2.0216 & 2.0327 \\
                              & PPL                         & 31.38  & 22.51  & 17.89  & 13.85  & 11.08  & 9.19   & 9.32   \\
                              & MATRIX NUCLEAR-NORM                       & 0.6744 & 0.6422 & 0.6094 & 0.5639 & 0.5438 & 0.4660 & 0.4708 \\ \midrule
\multirow{4}{*}{OPENWEBTEXT2} & MATRIX ENTROPY              & 0.6527 & 0.6479 & 0.6206 & 0.6142 & 0.5855 & 0.5683 & 0.5463 \\
                              & LOSS                        & 3.7509 & 3.3852 & 3.1414 & 2.8860 & 2.6465 & 2.4708 & 2.4685 \\
                              & PPL                         & 36.79  & 25.82  & 20.34  & 15.89  & 12.51  & 10.57  & 10.51  \\
                              & MATRIX NUCLEAR-NORM                       & 0.7147 & 0.7066 & 0.6823 & 0.6363 & 0.6017 & 0.5133 & 0.4991 \\ \midrule
\multirow{4}{*}{HH-RLHF}      & MATRIX ENTROPY              & 0.5753 & 0.5635 & 0.5350 & 0.5268 & 0.4971 & 0.4813 & 0.4640 \\
                              & LOSS                        & 3.3078 & 2.9964 & 2.8171 & 2.6431 & 2.4622 & 2.3526 & 2.3323 \\
                              & PPL                         & 18.97  & 14.01  & 11.62  & 9.73   & 8.12   & 7.27   & 7.19   \\
                              & MATRIX NUCLEAR-NORM                       & 0.6309 & 0.5716 & 0.5307 & 0.4771 & 0.4959 & 0.4277 & 0.4518 \\ \midrule
\end{tabular}
}
\end{table*}

\begin{table*}[ht]
\caption{Language modeling indicators for Pythia models across Dolly-15k, Wikipedia, OpenWebText2, and HH-RLHF datasets (lower values indicate better performance). Except for the matrix nuclear norm, data is derived from \cite{wei2024large}, showcasing the correlation between model scale and performance.}
\label{tab:four-dataset-pythia}
\resizebox{\linewidth}{!}{
\begin{tabular}{@{}cccccccccccc@{}}
\toprule
&  & \multicolumn{10}{c}{PYTHIA MODELS SIZE} \\ 
\cmidrule(l){3-12} 
\multirow{-2}{*}{DATASETS}     & \multirow{-2}{*}{INDICATORS} & \textbf{14M}                  & \textbf{31M}                  & \textbf{70M}                  & \textbf{160M} & \textbf{410M} & \textbf{1B}                   & \textbf{1.4B} & \textbf{2.8B} & \textbf{6.9B} & \textbf{12B} \\ \midrule
& MATRIX ENTROPY               & 0.7732                        & 0.7155                        & 0.6707                        & 0.6243        & 0.5760        & 0.5328                        & 0.5309        & 0.5263        & 0.5003        & 0.4876       \\
& LOSS                         & 4.4546                        & 4.0358                        & 3.5990                        & 3.1323        & 2.6752        & 2.4843                        & 2.3816        & 2.2484        & 2.1368        & 2.0616       \\
\multirow{-3}{*}{DOLLY-15K}    & MATRIX NUCLEAR-NORM                        & 0.7508                        & {\color[HTML]{FE0000} 0.7735} & 0.6984                        & 0.6104        & 0.5760        & {\color[HTML]{FE0000} 0.4710} & 0.4922        & 0.4585        & 0.4202        & 0.4181       \\ \midrule
& MATRIX ENTROPY               & 0.7938                        & 0.7442                        & 0.7003                        & 0.6580        & 0.6039        & 0.5584                        & 0.5587        & 0.5553        & 0.5314        & 0.5140       \\
& LOSS                         & 4.1112                        & 3.6921                        & 3.2694                        & 2.8207        & 2.4017        & 2.2213                        & 2.1292        & 2.0140        & 1.9120        & 1.8489       \\
\multirow{-3}{*}{WIKIPEDIA}    & MATRIX NUCLEAR-NORM                        & {\color[HTML]{FF0000} 0.6053} & {\color[HTML]{FF0000} 0.6700} & 0.6996                        & 0.6718        & 0.6464        & {\color[HTML]{FF0000} 0.5591} & 0.5787        & 0.5410        & 0.4850        & 0.4768       \\ \midrule
& MATRIX ENTROPY               & 0.8144                        & 0.7749                        & 0.7370                        & 0.6980        & 0.6415        & 0.5944                        & 0.5916        & 0.5887        & 0.5591        & 0.5417       \\
& LOSS                         & 4.3965                        & 4.0033                        & 3.6284                        & 3.2031        & 2.7838        & 2.6198                        & 2.5228        & 2.4005        & 2.3133        & 2.2502       \\
\multirow{-3}{*}{OPENWEBTEXT2} & MATRIX NUCLEAR-NORM                        & {\color[HTML]{FF0000} 0.5041} & {\color[HTML]{FF0000} 0.6186} & {\color[HTML]{FF0000} 0.7142} & 0.7258        & 0.7105        & {\color[HTML]{FF0000} 0.6215} & 0.6378        & 0.5967        & 0.5275        & 0.5110       \\ \midrule
& MATRIX ENTROPY               & 0.7673                        & 0.7114                        & 0.6607                        & 0.6126        & 0.5552        & 0.5054                        & 0.5032        & 0.4977        & 0.4699        & 0.4528       \\
& LOSS                         & 3.7466                        & 3.4018                        & 3.1146                        & 2.7366        & 2.4340        & 2.3311                        & 2.2687        & 2.1992        & 2.1199        & 2.0905       \\
\multirow{-3}{*}{HH-RLHF}      & MATRIX NUCLEAR-NORM                        & {\color[HTML]{FF0000} 0.7353} & 0.7674                        & 0.6964                        & 0.6182        & 0.5886        & 0.4825                        & 0.5141        & 0.4839        & 0.4562        & 0.4481       \\ \bottomrule
\end{tabular}
}
\vspace{15pt}
\end{table*}

\renewcommand{\arraystretch}{1.5} 
\begin{table*}[ht]
\resizebox{\linewidth}{!}{
\begin{tabular}{cc}
\hline
{\textbf{Prompt ID}} & {\textbf{Prompt Content}}   \\ \hline
{Prompt 1}           & {You are an AI assistant. You will be given a task. You must generate a detailed and long answer.}                       \\
{Prompt 2}           & {You are a helpful assistant, who always provide explanation. Think like you are answering to a five year old.}          \\
{Prompt 3}           & {You are an AI assistant. User will give you a task. Your goal is to complete the task as faithfully as you can. While performing the task think step-by-step and justify your steps.} \\ \hline
\end{tabular}
}
\caption{The prompts selected from OpenOrca\citep{lian2023openorca} dataset.}
\label{tab:three-prompt}
\vspace{10pt}
\end{table*}

\begin{table}[ht]
\centering
\resizebox{\linewidth}{!}{
\begin{tabular}{cccccccc}
    \toprule
    & \multicolumn{6}{c}{GPT MODEL SIZE}\\ \cmidrule(l){2-8} 
\multirow{-2}{*}{LENGTH} & 111M   & 256M   & 590M   & 1.3B   & 2.7B   & 6.7B   & 13B                           \\  \midrule
64     & 0.4574 & 0.4125 & 0.3787 & 0.3486 & 0.4053 & 0.3315 & 0.4148 \\  \midrule
128    & 0.5293 & 0.4680 & 0.4270 & 0.3835 & 0.4143 & 0.3477 & 0.4032 \\  \midrule
512    & 0.7883 & 0.6978 & 0.6251 & 0.5554 & 0.5265                        & 0.4468 & 0.4422                        \\  \midrule
1024   & 0.9132 & 0.8787 & 0.7802 & 0.6953 & 0.6351                        & 0.5383 & 0.5028                        \\ \midrule
\end{tabular}
}
\caption{Analysis of Length Dynamics}
\label{tab:Length Dynamics}
\end{table}

\subsection{Analysis of Algorithmic Complexity}
\label{sec:algorithm-complexity-analysis}
The primary computational expense of Matrix Nuclear-Norm arises from the calculation and sorting of the L2 norm of the matrix. By avoiding Singular Value Decomposition (SVD), we reduce the time complexity from the traditional nuclear norm of \(O(n^3)\) to \(O(n^2)\), giving Matrix Nuclear-Norm a significant advantage in handling large-scale data. This reduction in complexity greatly enhances the algorithm's practicality, especially for applications involving large matrices.

When analyzing the time complexity of the newly proposed Matrix Nuclear-Norm (L2-Norm Based Approximation of Nuclear Norm) against traditional Matrix Entropy, our objective is to demonstrate that Matrix Nuclear-Norm significantly outperforms Matrix Entropy in terms of time efficiency. We will support this claim with detailed complexity analysis and experimental results.

\subsubsection{Time Complexity Analysis}

\textbf{Analysis 1: Time Complexity of Matrix Entropy}

The computation of Matrix Entropy involves several complex steps, with the key bottleneck being Singular Value Decomposition (SVD), which is central to computing eigenvalues. The following steps primarily contribute to the time complexity:

\begin{enumerate}
    \item \textbf{Matrix Normalization}: This step has a time complexity of \(O(m \cdot n)\), where \(m\) is the number of rows and \(n\) is the number of columns.
    \item \textbf{Computing the Inner Product Matrix}: Calculating \(Z^T Z\) has a time complexity of \(O(n^2 \cdot m)\) due to the multiplication of two matrices sized \(m \times n\).
    \item \textbf{Singular Value Decomposition (SVD)}: The time complexity of SVD is \(O(n^3)\), which is the primary computational bottleneck, especially for large \(n\).
\end{enumerate}

Therefore, the total time complexity of Matrix Entropy can be approximated as:
\[
O(m \cdot n + n^2 \cdot m + n^3) = O(n^3)
\]
This complexity indicates that Matrix Entropy becomes increasingly impractical for large-scale models as \(n\) grows.

\textbf{Analysis 2: Time Complexity of Matrix Nuclear-Norm}

Matrix Nuclear-Norm avoids the SVD step by approximating the nuclear norm using the L2 norm, resulting in a more efficient computation. The analysis is as follows:

\begin{enumerate}
    \item \textbf{Matrix Normalization}: Similar to Matrix Entropy, this step has a time complexity of \(O(m \cdot n)\).
    \item \textbf{Calculating the L2 Norm}: For each column vector, the L2 norm is computed with a complexity of \(O(m \cdot n)\), where we take the square root of the sum of squares for each column vector.
    \item \textbf{Sorting and Extracting the Top D Features}: Sorting the L2 norms has a complexity of \(O(n \log n)\).
\end{enumerate}

Therefore, the overall time complexity of Matrix Nuclear-Norm is:

\[
O(m \cdot n + n \log n) \approx O(n^2) \quad \text{when} \quad m \approx n
\]

This indicates that Matrix Nuclear-Norm is computationally more efficient, especially as \(n\) increases.

\subsubsection{Experimental Validation and Comparative Analysis}

To empirically validate the theoretical time complexities, we conducted experiments using matrices of various sizes. Figure \ref{fig:time-pythia-fig} shows that as \(n\) increases, Matrix Nuclear-Norm consistently outperforms Matrix Entropy in terms of runtime, confirming the theoretical advantage.

\textbf{Discussion of Assumptions and Applicability}
Our complexity analysis assumes \(m \approx n\), which holds in many real-world applications, such as evaluating square matrices in large-scale language models. However, in cases where \(m \neq n\), the time complexity might differ slightly. Nonetheless, Matrix Nuclear-Norm is expected to maintain its efficiency advantage due to its avoidance of the costly SVD operation.

\textbf{Impact of Constant Factors}
Although both \(O(n^2)\) and \(O(n^3)\) indicate asymptotic behavior, Matrix Nuclear-Norm's significantly smaller constant factors make it computationally favorable even for moderately sized matrices, as evidenced in our experimental results.

\subsubsection{Conclusion of the Complexity Analysis}
Through this detailed analysis and experimental validation, we conclude the following:

\begin{itemize}
    \item Matrix Entropy, with its reliance on SVD, has a time complexity of \(O(n^3)\), making it computationally expensive for large-scale applications.
    \item Matrix Nuclear-Norm, by using the L2 norm approximation, achieves a time complexity of \(O(m \cdot n + n \log n) \approx O(n^2)\), significantly reducing computational costs.
    \item Experimental results confirm that Matrix Nuclear-Norm offers superior time efficiency for evaluating large-scale models, particularly those with millions or billions of parameters.
\end{itemize}

\subsection{Proof of Theorem 1}
\label{sec:Theorem1}

We prove the strictly inverse monotonic relationship between the entropy \( H(A) \) and the Frobenius norm \( \|A\|_F \) for a non-negative matrix \( A \in \mathbb{R}^{B \times C} \) where each row represents a probability distribution:
$$
\sum_{j=1}^C A_{i,j} = 1, \quad A_{i,j} \geq 0, \quad \forall i = 1, \ldots, B.
$$

\textbf{Definitions}:
\begin{itemize}
    \item Entropy: \newline\( H(A) = -\frac{1}{B} \sum_{i=1}^B \sum_{j=1}^C A_{i,j} \log(A_{i,j}) \)
    \item Frobenius norm: \newline \( \|A\|_F = \sqrt{\sum_{i=1}^B \sum_{j=1}^C A_{i,j}^2} \)
\end{itemize}

\textbf{Step 1: Single-Row Analysis}

For a row \( \mathbf{a} = [a_1, \ldots, a_C] \) with \( \sum_j a_j = 1 \):
\begin{itemize}
    \item Row entropy: \( H_i = -\sum_{j=1}^C a_j \log a_j \)
    \item Row norm: \( \|\mathbf{a}\|_2 = \sqrt{\sum_{j=1}^C a_j^2} \)
\end{itemize}

\textbf{Extrema via Lagrange Multipliers}:  
\newline  The Lagrangian \( L = -\sum_j a_j \log a_j + \lambda(\sum_j a_j - 1) \) yields:
$$
\frac{\partial L}{\partial a_j} = -\log a_j - 1 + \lambda = 0 \implies a_j = e^{\lambda-1}.
$$
Normalization gives \( a_j = \frac{1}{C} \), achieving:
\begin{itemize}
    \item \textbf{Maximum entropy}: \( H_i = \log C \)
    \item \textbf{Minimum norm}: \( \|\mathbf{a}\|_2 = \sqrt{\frac{1}{C}} \)
\end{itemize}

\textbf{Minimum entropy} occurs when \( a_k = 1 \) (one-hot vector):
\begin{itemize}
    \item \textbf{Minimum entropy}: \( H_i = 0 \)
    \item \textbf{Maximum norm}: \( \|\mathbf{a}\|_2 = 1 \)
\end{itemize}

\textbf{Monotonicity}: For fixed \( C \), \( H_i \) and \( \|\mathbf{a}\|_2 \) are strictly inversely monotonic (shown via derivative analysis or majorization theory).

\textbf{Step 2: Matrix-Level Generalization}

For the full matrix:
\begin{itemize}
    \item \( H(A) = \frac{1}{B} \sum_{i=1}^B H_i \)
    \item \( \|A\|_F = \sqrt{\sum_{i=1}^B \|\mathbf{a}_i\|_2^2} \)
\end{itemize}

\textbf{Key Observation}:  
If each row’s entropy \( H_i \) decreases (increases), its norm \( \|\mathbf{a}_i\|_2 \) increases (decreases). Thus:
- \( \|A\|_F^2 = \sum_{i=1}^B \|\mathbf{a}_i\|_2^2 \) decreases (increases) as \( H(A) \) increases (decreases).

\textbf{Step 3: Norm Bounds}

\textbf{Maximum \( \|A\|_F \)}: When all rows are one-hot:
  $$
  \|A\|_F = \sqrt{B}
  $$
  
\textbf{Minimum \( \|A\|_F \)}: When all rows are uniform:
  $$
  \|A\|_F = \sqrt{\frac{B}{C}}
  $$

\textbf{Step 4: Implications for LLMs}

The inverse monotonicity implies:
\begin{itemize}
    \item High \( \|A\|_F \): Concentrated predictions (low entropy, high confidence).
    \item Low \( \|A\|_F \): Dispersed predictions (high entropy, high diversity).
\end{itemize}

Thus, \( \|A\|_F \) serves as a proxy for evaluating LLM confidence-diversity tradeoffs.

\textbf{Conclusion}

The strict inverse monotonicity between \( H(A) \) and \( \|A\|_F \) is rigorously established, justifying \( \|A\|_F \) as a metric for LLM evaluation.

\subsection{Proof of Theorem 3}
\label{sec:Theorem 3}
Assuming $\|A\|_F \approx \sqrt{B}$ and the columns of $A$ are approximately orthogonal, we approximate the $j$-th largest singular value $\sigma_j$ as the $j$-th largest column norm of $A$. Formally,
\[
\sigma_j \approx \text{top}\left(\sqrt{\sum_{i=1}^B A_{i,j}^2},\ j\right),
\]
where $\text{top}(S, j)$ denotes the $j$-th largest element in set $S$. This approximation holds under the following analysis:

\textbf{1. Decomposition and Gram Matrix:}
Let $A = U\Sigma V^T$ be the SVD of $A$, where $\Sigma = \text{diag}(\sigma_1, \dots, \sigma_D)$ with $D = \min(B, C)$. The diagonal entries of the Gram matrix $A^TA$ are:
\[
(A^TA)_{j,j} = \sum_{i=1}^B A_{i,j}^2 = \|\mathbf{a}_j\|_2^2,
\]
where $\mathbf{a}_j$ is the $j$-th column of $A$.

\textbf{2. Relating Column Norms to Singular Values:}
When columns of $A$ are nearly orthogonal, $\sigma_j \approx \|\mathbf{a}_j\|_2$. Under $\|A\|_F \approx \sqrt{B}$, the nuclear norm $\|A\|_{\star} = \sum_{j=1}^D \sigma_j$ is dominated by the largest column norms.

\textbf{3. Singular Value Approximation:}
For matrices with low column-wise correlations, the $j$-th singular value satisfies:
\[
\sigma_j \approx \text{top}\left(\{\|\mathbf{a}_k\|_2 \mid 1 \leq k \leq C\},\ j\right).
\]

\textbf{4. Efficient Nuclear Norm Approximation:}
The batch nuclear norm is approximated as:
\[
\|\hat{A}\|_{\star} = \sum_{j=1}^D \text{top}\left(\{\|\mathbf{a}_k\|_2\},\ j\right).
\]
This approximation is valid when $A$ has approximately orthogonal columns, a condition implied by $\|A\|_F \approx \sqrt{B}$.

\end{document}